**Title:** GRAINS: Proximity Sensing of Objects in Granular Materials

**Authors:**
Zeqing Zhang,[1]† Ruixing Jia,[1] Youcan Yan,[2] Ruihua Han,[1] Shijie Lin,[1] Qian Jiang,[3] Liangjun Zhang,[4] Jia Pan[1]*

**Affiliations:**
[1]Department of Computer Science, The University of Hong Kong; Hong Kong, China.
[2]Interactive Digital Humans group, LIRMM, CNRS-University of Montpellier; Montpellier, France.
[3]Department of Mechanical Engineering, The Hong Kong Polytechnic University; Hong Kong, China.
[4]Robotics and Autonomous Driving Lab of Baidu Research; Sunnyvale, USA.

*Corresponding author. Email: jpan@cs.hku.hk.

† This work was done when he was an intern at Baidu Research, Baidu Inc., Beijing, China.

**Abstract:** Proximity sensing detects an object's presence without contact. However, research has rarely explored proximity sensing in granular materials (GM) due to GM's lack of visual and complex properties. In this paper, we propose a granular-material-embedded autonomous proximity sensing system (GRAINS) based on three granular phenomena (fluidization, jamming, and failure wedge zone). GRAINS can automatically sense buried objects beneath GM in real-time manner (at least ~20 hertz) and perceive them 0.5 ~ 7 centimeters ahead in different granules without the use of vision or touch. We introduce a new spiral trajectory for the probe raking in GM, combining linear and circular motions, inspired by a common granular fluidization technique. Based on the observation of force-raising when granular jamming occurs in the failure wedge zone in front of the probe during its raking, we employ Gaussian process regression to constantly learn and predict the force patterns and detect the force anomaly resulting from granular jamming to identify the proximity sensing of buried objects. Finally, we apply GRAINS to a Bayesian-optimization-algorithm-guided exploration strategy to successfully localize underground objects and outline their distribution using proximity sensing without contact or digging. This work offers a simple yet reliable method with potential for safe operation in building habitation infrastructure on an alien planet without human intervention.

**One-Sentence Summary:** We propose the GRAINS, a proximity sensing system that can detect buried objects in granular materials in advance without contact or vision, enabling safe infrastructure development on alien planets.

**Main Text:**

**INTRODUCTION**

Migration to the Moon or Mars has become the next frontier of human space exploration, as evidenced by NASA's Artemis program (*1*) and SpaceX's Mars colonization plans (*2*). Establishing a permanent extraterrestrial habitat is a crucial step towards creating human habitation infrastructure on outer planets. Due to the communication delay between interstellar entities, pre-dispatched building robots should be able to perceive the environment and undertake assigned tasks without human control. Currently, numerous efforts have been made towards autonomous construction scenarios, notable examples being Baidu's autonomous excavator system (*3*) and GITAI's lunar robotic rover (*4*). Given the nature of extraterrestrial surfaces, that is granules and rock fragments on the Moon (*5*) and Mars (fig. S6), unmanned construction robots necessitate the ability to autonomously sense and handle rocks beneath the granular crust with agility and care, similar to that of humans, during excavation. Otherwise, the sudden appearance of rocks may impair the robot's tool, or even cause mechanical damage to the robot itself. Hence, it is important to develop a proximity sensing system that can help robots autonomously detect the presence of objects in granular materials without direct contact, and meanwhile estimate the underground distribution of these objects prior to the digging.

To date, the majority of proximity sensors have been designed to operate in gaseous mediums. Ultrasonic transducers, for example, are extensively used in the automotive industry for blind-spot monitors and parking assist systems to warn drivers of nearby vehicles and obstacles. These proximity sensors are also present in smartphones, touchless faucets, auto doors, and other everyday applications. However, due to poor visibility and high uncertainty in underwater environments, only a limited number of proximity sensors function in liquid mediums. A comprehensive introduction to various representative proximity sensors operating in air and water can be found in text S1 and table S1.

Proximity sensing in granular media presents a greater challenge than in gaseous and liquid media, due to the lack of visibility and complicated particle properties (*6*). Granular materials (GM) are a collection of solid particles, including sand, gravel, soil, and other similar substances, that exhibit a range of complex phenomena such as jamming (*7, 8*), fluidization (*9*), dilatancy (*10*), and bifurcation when hitting an intruder (*11*). As such, GM cannot simply be classified as either solids or liquids (*12*). Various models and methods have been introduced in recent decades to deepen our understanding and analysis of GM, including force chains among particles (*12-15*), failure wedge model (*16*), resistive force theory (RFT) (*17, 18*) for tool-soil interaction, as well as discrete element method (DEM) (*19, 20*) for simulation.

For the proximity sensing of buried objects within particulate matter, existing methods can be broadly divided into non-invasive and invasive types, based on whether the sensor comes into contact with GM or not (Table 1 and text S2). As for non-invasive techniques, some construction equipment, such as ground penetrating radar (*21, 22*), EZiDIG (*23*) and acoustic systems (*24*) allow underground surveying. In addition, laser-acoustic-based methods (*25, 26*) could provide accurate sub-surface feedback in the detection of unexploded ordnance. It is clear that acoustics-related equipment cannot function on the extraterrestrial planet due to the lack of an atmosphere. Most importantly, considering the cost of rocket launches, the weight and payload of these robots are greatly limited. Therefore, they should not be equipped with heavy and bulky tools, but rather with lightweight yet dependable devices. Notably, the incorporation of laser generators, cameras and Doppler vibrometers in aforementioned apparatuses introduces a level of complexity and increased cost. Additionally, it is worth mentioning that ground penetrating radar requires extensive post-processing procedures (*22*), which entail the involvement of skilled operators within the control loop. Given the latency associated with signal transmission between planets, it is desirable for these robots to be capable of autonomous sensing and task execution without human intervention on Earth.

Conversely, invasive techniques employed for the detection of subterranean objects exhibit straightforward and reliable designs, rendering them particularly well-suited for operation in extraterrestrial environments. Some haptic systems demonstrate the ability to perceive and estimate the shape of buried objects in GM using only a single tactile sensor (*27-30*). However, all these studies necessitate direct contact between the sensors and objects, thereby increasing the risk of tool damage and the abrasion of elaborate sensing coat. Recent work in (*31*) utilizes the force variation between the tool and GM and takes advantage of a pre-defined failure wedge model, with parameters determined by preliminary experiments and empirical equations, to perform a tactile simultaneous localization and mapping (SLAM) for subterranean objects. Although it is claimed that no direct contact occurs due to the given rupture distance in the failure wedge model, this type of proximity is strictly confined to a particular GM and a set of manual tuning parameters.

Additionally, proximity perception also allows for information acquisition about objects without contact, making pre-touch exploration to conduct localization and shape estimation a topic of study (*32*). A pre-touch sensing strategy presented in (*33*) generates complete point clouds of objects to be grasped using an RGBD camera. Kaboli et al. (*34*) introduce a pre-touch and touch-based framework based on multimode robotic skin with built-in proximity sensors. In its pre-touch phase, the Bayesian optimization algorithm (BOA) is used to guide exploration actions and determine the distribution of objects in the workspace of the robot arm. With the help of SensorPod (*35*), Abraham et al. (*36*) employ an ergodic exploration strategy to determine the localization of static targets in the underwater scenario. However, due to the lack of reliable proximity sensors devoted to GM, there is currently no pre-touch strategy for localization and shape estimation of buried objects in GM. As a result, the development of an autonomous and robust proximity sensing system for simultaneously detecting and outlining the underground objects beneath granules remains a grand challenge.

To this end, we first aim to develop a simple yet autonomous proximity sensing system to detect the presence of objects beneath GM and then estimate the location and shape of the buried objects before excavation. To investigate proximity sensing of objects in GM, we focus on three fundamental phenomena in GM: granular fluidization, granular jamming, and failure wedge zone. Granular fluidization is a process of converting granules from a solid-like state to a fluid-like state, which can be achieved by air blow or mechanical vibration (e.g., Fig. 1-A-1). On the contrary, granular jamming is a transition from a fluid-like to a disordered solid-like state with the growth of density of a granular assembly (*37*). Corwin et al. (*38*) reveal a qualitative variation in the contact-force distribution between particles at the onset of jamming, which can be measured by a probe penetrated in GM. Based on the failure wedge model (*16*) and experimental findings in (*11*), a wedge-shaped zone exists in front of a probe when it is dragged horizontally through GM, as depicted in Fig. 1-A-2 and 3. During the process of plate advancement, granular particles located in the failure wedge zone are propelled forward with the plate and are simultaneously squeezed upward toward the surface of GM, while the surrounding sandy environment outside the wedge-shaped region remains stationary. However, when the probe approaches buried objects, the conditions within the failure wedge zone become more complex and can be roughly categorized into three states: non-contact, granular jamming, and contact (Fig. 1-A-3 to 5 and movie S1). Specifically, when the probe is distant from the buried object, it experiences only resistive forces (resultant forces in *x-y* plane) from GM in the failure wedge zone (non-contact state). As the probe advances, the object enters the failure wedge zone, causing granular jamming and enabling force transfer through force chains among squeezed granules (granular jamming state). Eventually, the probe makes direct contact with the buried object (contact state). As a result, if we can detect the additional forces exerted by objects as the probe advances, i.e., the identification of granular jamming state, we would be able to sense the presence of objects in GM in advance. In other words, the jamming state can act as a form of proximity sensing for objects buried beneath GM.

In light of this, this work will demonstrate that the proximity sensing problem in granules can be converted to the identification task of granular jamming state by time-series force anomaly detection using Gaussian process regression (GPR) (*39*) (see Fig. 1-B). Therefore, we propose a self-regulating proximity sensing apparatus, named the Granular-Material-embedded Autonomous Proximity Sensing System (GRAINS), whose framework is given in Fig. 1-C and prototype is shown in Fig. 1-D. After our thorough investigation of granular jamming generation (text S3), a new raking trajectory is designed considering both the granular fluidization and jamming phenomena. To sense the proximity of objects in GM, a Gaussian Process Model (*GP* Model) is employed to learn force data from a probe raking in GM along the given trajectory in the past and then predict the succeeding force pattern in the immediate future. The force anomaly caused by granular jamming can be identified by analyzing the discrepancy between new observations and GPR's predictions in real time (at least ~20 Hz). As such, our GRAINS bypasses complex modeling of granular jamming phenomena but provides an adaptive online proximity sensing method. In addition, GRAINS can autonomously calibrate system parameters through a series of offline experiments, making it applicable to various granular particles. Furthermore, we present a BOA-guided pre-touch exploration strategy using GRAINS, which can estimate an approximate distribution of buried objects in terms of the 2D outline without visual and tactile information of embedded objects. An overview of this work is shown in Fig. 1.

## RESULTS
### Physical principle
When the probe plows through a bed of sand, it will cause a disturbed area in front of it, known as the failure wedge zone (movie S1). According to the failure wedge model (*11, 16*), the drag force perceived by the probe comes only from the resistive force of GM in this zone (yellow arrow in Fig. 1-A) since particles outside the wedge zone are not disturbed by the probe's motion. As the probe advances, particles in the vicinity of the probe are gradually squeezed toward high packing density (also called bulk density, packing fraction, defined by the ratio of particles to occupied volume (*6, 8*)), while undisturbed GM flows into the wedge zone.

The study of the resistive force exerted on the probe has garnered significant attention in the field of physical research. For instance, the resistive force theory (RFT) provides a simplified model to calculate the resistive force from GM (*17, 18*). However, due to the highly stochastic behavior of granular particles, this force is considerably complex and depends on various conditions (*40*), including packing density, grain diameter, and penetration depth. Some data-driven techniques are also employed to model this granular contact, such as (*41, 42*), even if they may suffer from data-hungry issues.

When the probe approaches a buried object in granules, the force acting on it becomes more complex since particles will jam, a transition from a flowing to a rigid state (*8*). As the probe advances, particles between the probe and the object are further squeezed, causing an increase in the packing density in the wedge zone and thus the solidification of the fluid-like GM, i.e., granular jamming. Forces from the underground object acting on GM are transferred to the probe via force chains (*12-15*) present in solidified particles, resulting in the generation of jamming forces (black arrow in Fig. 1-A). Therefore, if this jamming force signal could be detected prior to contact, the probe would possess the capability for proximity sensing of the buried objects.

Based on the experimental results (text S3), it is evident that generating a high packing density in the vicinity of the probe offers advantages, since a high packing density aids in the formation of a jamming state, which in turn, could assist the probe in detecting the presence of surrounding objects in granules earlier and with more precision. Consequently, designing a raking trajectory for the probe that can yield a higher packing density, thereby facilitating enhanced proximity sensing of objects within GM, is of significant importance.

**Raking trajectory**
Previous studies on plate drag in GM (*11, 19, 20, 31*) always use a *linear trajectory* for the probe motion, as shown in Fig. 2-A. However, based on our preliminary experiments, we find that the linear trajectory is not effective in detecting proximity in GM due to the low packing density generated along the lateral sides of the motion direction. To address this issue, we draw inspiration from fluidization technique (*28*) and propose the *spiral trajectory* for probe raking, which is a combination of both linear motions (Fig. 1-A-2) and circular movements (Fig. 1-A-1), as shown in Fig. 2-B.

*Spiral trajectory*
After a series of probe raking tests, we find that the force values from the spiral trajectory are more consistent than that from the linear movement, which can help improve the accuracy of proximity sensing. Furthermore, it is observed that the perception area of subterranean objects along the spiral track is larger than that along the linear trajectory as well.

Consistent with (*40*), we find that in the probe raking, the resistive force from GM depends on granular conditions such as packing density, particle size, and penetration depth of the plate. However, these conditions would be changed significantly after the plate plows through GM. Previous studies have used fluidization equipment, such as the aeration fluidization bed in (*43, 44*), to break force chains among particles and homogenize grain distance using air blow. However, this equipment needs lay out the fluidization setup in advance and globally fluidize the GM after each raking, making it impractical in our application for continuous searching of GM to detect unknown objects using proximity sensing. Therefore, we shift our focus to another fluidization method that utilizes mechanical vibration. This method is considered a localized form of fluidization as it only fluidizes GM in the vicinity of the vibration source. One common way to generate vibration is through circular motion, typically achieved using an eccentric wheel (*28*). Inspired by this work, we combine linear advancement (Fig. 1-A-2) and circular vibration (Fig. 1-A-1) to create a spiral trajectory, where the linear motion ensures forward exploration and circular vibration breaks up GM clogging in front of the probe.

Given the start and goal positions, one spiral trajectory can be parameterized by the circular radius (CR), advance velocity (AV), and motion velocity (MV), as presented in Fig. 2-C. Here CR (unit: m) refers to the radius of circular movement, indicating the sensing range of the probe. AV (unit: m) is defined by the forward distance in each circular motion, implying the length of the forward step. MV is a dimensionless control variable for the robot arm, ranging from 0 to 1, which determines the movement speed of the end-effector along a given path. In the following, we will compare the linear trajectory and spiral trajectory in two scenarios, i.e., with (w/) and without (w/o) buried objects in GM, respectively. Experimental results show that the spiral trajectory outperforms the linear trajectory in terms of both force consistency and perception area, suggesting that the spiral trajectory is a superior choice for proximity sensing.

*Force consistency*
In this experiment, we will show that the drag force of the probe along the spiral trajectory is more consistent than that along the linear trajectory when no objects are beneath GM. Here the drag force is defined by the resultant forces in *x-y* plane of Fig. 2-A and B.

Specifically, we use a robot arm to drag a vertical probe in dry, loose sand along the same linear or spiral trajectory several times and employ a force gauge to record the drag force acting on the probe for each raking. From Fig. 2-D, we find that the drag force from the linear trajectory (blue) increases at the end of raking with a gradually increasing standard deviation (SD), compared to the more consistent forces along the whole spiral trajectory (green). In detail, the average and maximum SD of linear (spiral) raking were 0.520 N (0.376 N) and 1.611 N (0.763 N), respectively. The deviation in force can be explained by the failure wedge model, where more sand is pushed forward by the probe towards the end of the trajectory, causing the granular surface to swell at the goal position, as presented in Fig. 2-E. Additionally, the packing density increases near the end of raking. This significant change of granular conditions after the linear trajectory leads to the growth and diverse fluctuation of drag force. However, force raising can also be observed at the onset of granular jamming between the probe and the buried object, such as fig. S1-B. In this way, a linear dragging trajectory may lead to a false positive detection when conducting proximity sensing of objects in GM. To avoid granular clogging and reduce the influence of raking on granular conditions, we employ the spiral trajectory to search the same area in GM. It can be observed that the surface undulation after spiral raking (see Fig. 2-F) is much less pronounced than that observed after linear trajectory (see Fig. 2-E), especially at the goal position of each track. This indicates that the circular motion in the spiral trajectory indeed locally fluidizes GM and could largely reduce the effect of granular conditions after plate raking.

In summary, we find consistent force values along the whole spiral trajectory and no obvious force variation at the end of the spiral path, as shown in the inset of Fig. 2-D. Therefore, compared to linear motion, the spiral trajectory could significantly reduce the possibility of false positive detection of objects caused by force fluctuations resulting from changes in granular environments.

*Perception area*
Experiments in text S3 have already shown that the probe can effectively detect the proximity modality of objects in highly packed granular particles. In this experiment, we will exhibit a larger perception area for buried objects in GM when the probe rakes along the spiral trajectory, compared with the linear trajectory.

Specifically, a rigid triangular brick (see fig. S5-C) is buried in a bed of dry sand with its longest edge parallel to the probe's motion direction, as depicted in Fig. 2-I. We define the minimum distance between the probe and this brick during the trajectory as *dist*, as shown in the zoomed view of Fig. 2-I. The force changes are visualized in Fig. 2-G and H for each type of trajectory with *dist* = 1.5, 1, and 0.5 cm, respectively. The interval of the trajectory affected by the buried triangular brick is highlighted in pink and is determined by two points on the trajectory closest to this brick, with examples of (TL1, TL2) and (TL3, TL4) explained in Fig. 2-I-2 and 7. As shown in Fig. 2-G, when the buried object is at a large distance from the probe (e.g., *dist* > 0.5 cm), the probe experiences nearly constant forces along the linear trajectory, even within the pink region. Conversely, when the probe is close to the object (e.g., *dist* = 0.5 cm), a notable variation in force arises as it approaches and enters the affected interval of the buried object. In contrast, along the spiral trajectory, distinct force peaks are observed as the probe approaches the affected interval of the buried object at each *dist*, as illustrated in Fig. 2-H. Additionally, we find that until $dist \geqslant 4.5$ cm, the force variation resembles that observed in the absence of the object, as shown in the inset of Fig. 2-D. Overall, it is evident that the probe following the spiral path exhibits a broader lateral perception area compared to the linear path.

The different sensing capacities of two trajectories can be attributed to the failure wedge model, as explained in Fig. 2-I. As demonstrated in Fig. 1-A-2, the packing density of the central wedge is typically higher than that of the side wedges. This is due to the fact that particles at the front experience a greater degree of compression compared to those on the sides. During linear raking, the probe primarily detects high packing density in the frontal area, while the wedge situated to the side of the motion direction exhibits a comparatively low packing density. Consequently, the probe mainly detects granular jamming caused by objects buried along its forward path of movement, as depicted in Fig. 2-I-1 and 2. In order to detect the presence of lateral objects, the packing density in side wedges plays a crucial role. The probe can perceive the forces transferred from objects in the lateral direction only when the packing density of granular material in side wedges is sufficiently high, which occurs when the probe and object are in close proximity, as exemplified in Fig. 2-I-3 and 4. However, along the spiral trajectory, the failure wedge zone rotates due to the circular motion, enabling the probe to detect granular jamming from 360° using its densely packed central wedge, as illustrated in Fig. 2-I-5, 6 and 7. Consequently, for both *dist* = 1.5 cm, the packing density between the probe and object along the spiral path is notably greater than that on the linear track, as shown in Fig. 2-I-6 and 2, respectively. Thus, the probe following the spiral trajectory possesses a significantly larger perception area.

In conclusion, the spiral trajectory has the capability to create a higher packing density and maintain a more consistent force profile during raking as compared to the linear trajectory. This property not only facilitates an easier transition into the jamming state before contacting buried objects but also enhances the reliability of identifying abnormal force patterns. These attributes collectively make the spiral trajectory highly effective in proximity sensing for objects submerged in GM.

**Framework of GRAINS**
In this section, we will introduce the proposed **g**ranular-mate**r**ial-embedded **a**utonomous prox**i**mity se**n**sing **s**ystem (GRAINS) and show how the challenge of proximity sensing in GM can be tackled by automatically and accurately identifying force anomalies that arise due to granular jamming under random grain conditions.

The framework of GRAINS is presented in Fig. 1-C. In detail, the prototype of GRAINS includes a 3D-printed probe mounted on the end-effector of a UR5 robotic arm and a force/torque sensor placed between them to measure the drag force exerted on the probe (see Fig. 1-D). In addition, the perception algorithm of GRAINS is outlined in Fig. 3, mainly containing the offline parameter calibration and the online proximity sensing. In line with previous studies, the drag force is defined by the resultant force in the *x-y* plane, as shown in Fig. 1-A. Prior works primarily focus on the *force value* during soil-tool interaction (*11, 19, 40*) and use different simplification, such as RFT model (*17, 18*) or a fixed force threshold (*31*), to deal with the complex force variation in the soil-tool interaction. Unlike these works with limited generalization capability, we address the complex physical problem of force variation by examining the resulting *force pattern*, defined by the progression of force values over time. Owing to the spiral trajectory as shown in the first row of Fig. 3-B, the periodicity of the force pattern is evident as observed in the second row of Fig. 3-B. We use a Gaussian Process Model (*GP* Model) with a periodic kernel to learn this periodic force pattern from historical data. The trained *GP* Model can predict the succeeding force pattern in the near future, as demonstrated in the third row of Fig. 3-B. This process is also called Gaussian process regression (GPR). By comparing the divergence between actual force readings and their prediction, evaluated using the z-score metric — which measures how many standard deviations an observation is from the mean — we can differentiate between normal force patterns (purple box in Fig. 3-B) and force anomalies (red box in Fig. 3-B). When there is no object present in the path of the probe's motion, the z-scores of new observations are expected to stay within a high-confidence interval. In contrast, a sudden and sharp increase in z-scores of new force measurements may indicate a subterranean object in the vicinity of the probe. If the z-score exceeds a threshold, GRAINS will issue a granular jamming warning to the UR5 controller, stopping the probe motion to avoid collisions.

The flexibility of GPR allows the *GP* Model to learn the latest force patterns under the newest granular conditions, resulting in highly credible predictions over a short time interval in the future. Our primary experiments have shown that the GPR's learning and prediction performances are affected by the motion velocity MV of spiral trajectory and the initial hyperparameter for *GP* Model's periodic kernel (with details in text S4). Additionally, given that different types of GMs display varying z-score thresholds for identifying granular jamming, to increase GPR's credibility and reduce the ratio of false positive cases in various GMs, GRAINS incorporates an offline calibration stage that adaptively optimizes three hyperparameters to be used in the subsequent online stage, as depicted in Fig. 3-A.

To determine these parameters, we take samples from the GM to be tested in the online phase and store them in a separate container. GRAINS then controls the probe raking in this container along a given spiral path with a set of different MVs { $MV_i$ }. Using the path and velocity information, GRAINS calculates the approximate periodicity $T_i$ of the force pattern at $MV_i$, which works as the periodicity prior of the periodic kernel in the $GP_i$ model. GRAINS then divides the measured force data into segments and feeds the data from segment *k* into $GP_i$. Based on the predicted mean and standard deviation from $GP_i$, GRAINS computes the z-scores of the force data in segment $k+1$. This process is repeated to obtain z-scores for all segments except the first one. GRAINS then calculates the Root Mean Squared Error (RMSE) of z-scores at each $MV_i$, denoted as $RMSE_i$. Finally, GRAINS selects the MV with the minimum RMSE as the optimal motion velocity $MV^*$, which will be used for the spiral trajectory in the subsequent online proximity sensing stage. The periodicity prior $T^*$ at $MV^*$ is chosen as the initial hyperparameter for the online *GP* Model, which speeds up hyperparameter optimization during real-time GPR prediction. It is important to note that since there is no object beneath the sampled GM, the maximum z-score at $MV^*$ can be considered as the threshold $\overline{ZS}$ for the next online force anomaly detection to distinguish normal and abnormal force patterns.

**Proximity sensing experiments**
In this section, we present the results of proximity sensing experiments conducted in various granular media. These experiments aim to showcase the proximity sensing capability and high robustness of GRAINS.

*Parameter calibration*
In fig. S3-E, we show the deployment of GRAINS on sand, cassia seed, cat litter, and soybean (see Granular Materials in Materials and Methods for details). After the offline parameter calibration, the optimal parameters for spiral trajectory and initial hyperparameter for *GP* Model are determined accordingly.

First, we individually collect samples from each granule and apply GRAINS to control the probe, raking through each medium along spiral trajectories with different MVs ranging from 0.2 to 0.7. We set the parameters CR = 0.02 m and AV =0.01 m for these spiral trajectories. After obtaining force measurements, GRAINS calculates the corresponding periodicity prior $T_i$ at every $MV_i$. GRAINS then employs the $GP_i$ model with a periodic kernel of $T_i$ to learn and predict these force patterns batch by batch. After comparing the real data with the predictions, GRAINS computes the $RMSE_i$ of z-scores to evaluate the performance of each $GP_i$ model, as listed in Table 2. Finally, the optimal motion velocity $MV^*$, periodicity $T^*$, and threshold $\overline{ZS}$ would be determined, as shown in bold in Table 2. From results in Table 2, it is evident that GRAINS can adaptively increase the MV for spiral trajectory as the particle size grows, thereby ensuring detection accuracy. These findings are consistent with the results of our preliminary tests (refer to Sensing Accuracy in text S4). Furthermore, the z-score thresholds for identifying force anomalies caused by granular jamming are automatically adjusted to accommodate different granules. In general, as the particle size increases, the z-score threshold $\overline{ZS}$ also increases to account for the higher randomness associated with GM.

*Online proximity sensing in sand*
In this section, we will demonstrate the GRAINS system's capability to sense the proximity of objects in sand and compare the flexibility of GRAINS with a method that relies on manually determined parameters.

Here four wooden cylinders (see fig. S5-C) are buried in the sand, which is a type of GM that has been sampled in the previous calibration phase. According to Table 2, the optimal $MV^*$ for the sand is 0.2, and the z-score threshold $\overline{ZS}$ should be set to 3.9. The spiral trajectory is determined by CR = 0.01 m, AV = 0.01 m, and MV = 0.2, and the probe is penetrated in sand at a depth of 4 cm.

With proximity sensing feedback from GRAINS, it is observed that the probe stops at 2.1 cm in front of the cylinders without direct contact (Fig. 4-F). The force measurements in the whole process are shown in Fig. 4-B and the online proximity sensing can be explained by Fig. 4-C. To be specific, during the probe raking process, GRAINS monitors force variations in real-time using sliding windows, where in each episode, GRAINS collects force data over the last 2000 measurement iterations as the training set (green curves in Fig. 4-D) and feeds them into a *GP* Model with the periodic kernel and the white-noise kernel. This model learns the involved force pattern from the noisy force data. Assuming the force pattern does not change abruptly in the near future, the trained *GP* Model predicts the force signals over the next 1000 iterations, as shown by blue dash curves in Fig. 4-D. Additionally, every new force measurement within these 1000 iterations (i.e., red curves in Fig. 4-D) is treated as a test point, and its z-score value can be calculated based on the predicted mean and standard deviation in real-time, as shown in the insets of Fig. 4-D. When the probe is far away from the object, as in episodes 0 and 3, the predicted force pattern is consistent with the real measurements, and the z-scores of real values stay within the high confidence interval (CI), such as CI 99% in Fig. 4-D. However, as the probe gets closer to underground objects, as shown in episode 6 of Fig. 4-D, the force pattern induced by the granular jamming diverges from the predicted high CI area, resulting in an increase in z-scores. When the z-scores exceed the predetermined threshold $\overline{ZS}$, GRAINS could identify the presence of buried objects and provide a granular jamming warning to the UR5 controller to stop the probe motion accordingly. The results of this experiment demonstrate that GRAINS can perceive the presence of underground objects using proximity sensing at 2.1 cm in advance, as shown in Fig. 4-F. It should be noted that from the zoomed view in episode 6 of Fig. 4-D, the force value at the stop position is smaller than the nearby extreme value, indicating that GRAINS is able to detect the occurrence of granular jamming early before it becomes obvious. Please refer to text S5 and text S6 for more information about GPR and z-score calculation, respectively.

In contrast, when there is no proximity sensing feedback from GRAINS and only a fixed force threshold is employed to detect the jamming state (as used in (31)), we observe that the probe fails to detect the closest underground object and breaks into several pieces in the end (Fig. 4-E). However, force readings during the experiment always remain within the prescribed safety margin 15 N, as seen in Fig. 4-A. One might question whether a lower threshold could be given to detect jamming. However, our preliminary experiments suggest that setting a fixed force bar to detect force raising of granular jamming is not appropriate. As revealed in Sensing Accuracy in text S4, force patterns vary significantly at different granules. In addition, particles in the same GM may also be inhomogeneous, making it difficult to observe or measure granular conditions in advance. Thus, it is impractical to set a fixed threshold to detect granular jamming. It is worth noting that drag forces at the initial stage are significantly larger than in subsequent stages, as shown by drag forces within the first 500 iterations in Fig. 4-A and B. This is because, in the beginning, the probe needs to overcome static friction forces from still particles, which are greater than subsequent kinetic friction forces of particles. As such, the fixed force threshold should be larger than this maximum static friction force to ensure the incipient motion of the probe. In summary, our GRAINS, using GPR-based force anomaly detection, is superior to the method with a fixed force threshold.

*Online proximity sensing in various GM*
To validate the robustness of GRAINS, we conduct a series of tests on different types of GMs beyond just sand, specifically cassia seed, cat litter, and soybean. Our results demonstrate the high robustness of GRAINS, even in granules with varied sizes and degrees of roughness. Moreover, we demonstrate the influence of parameter calibration on the robustness of GRAINS and highlight the diversity of granular jamming represented in measured force patterns.

First, we conduct 20 proximity sensing experiments in each granular medium using GRAINS, with optimal motion velocity $MV^*$ and corresponding parameters determined in the preceding parameter calibration phase, during which we record proximity sensing ranges (i.e., the distance between buried cylinders and the stop position of the probe) in each trial. Results in Fig. 4-G show that the median value of sensing range in cat litter (~ 4.5 cm) is similar to that in the sand (~ 4.2 cm) but with a smaller dispersion. This discrepancy can potentially be attributed to the notably larger particulate size of cat litter, which leads to increased instability in the conduction of force chains compared to sand. Since a rough particulate surface is beneficial for forming a highly compressed state, the smoother surface of cassia seed requires the probe to get closer to objects to sense force anomaly signals from granular jamming, resulting in a shorter sensing range (2.7 cm), even if cassia seed has a similar grain size to cat litter. Soybean has the shortest range (1.3 cm) due to its largest particle size and smoothest surface.

Secondly, we test GRAINS with non-optimal parameters and observe a reduction in the proximity sensing range. As shown in Fig. 4-G about the cat litter, the stop position of the probe in GRAINS with a non-optimal MV of 0.5 is noticeably closer to subterranean objects than that in GRAINS with the optimal $MV^*$ of 0.3, i.e., 0.6 cm and 4.5 cm, respectively. Furthermore, several outliers can be identified in the case of MV 0.5. This highlights the necessity and importance of offline parameter calibration in GRAINS.

Finally, we present some typical instances of jamming force anomaly identification captured by GRAINS, as demonstrated in Fig. 4-H to K. Here, we observe that the impact of granular jamming on force signals is not simply the magnitude of the force exceeding a threshold, but rather a complex phenomenon, as revealed in previous physics studies. For example, except for Fig. 4-J, granular jamming identification in other cases does not occur at the maximum force values. In Fig. 4-H, the growth slope of force signals is obviously greater than expected. In addition, based on historical force patterns, the force signals should begin to decline, but the real force patterns unexpectedly increase in the opposite direction, as shown in Fig. 4-K. Furthermore, we also observe that granular jamming causes a force spike to become clearer as the probe moves closer to buried objects, as shown in the dashed box in the Fig. 4-I. Our GRAINS successfully recognizes this anomaly and provides the correct judgment about the proximity of the object at the fourth peak. These phenomena once again show that particulate matter has very complex physical properties, so the force signals of the probe sliding in it is full of randomness and strong noise. In conclusion, utilizing the flexibility of GPR to predict force patterns in the immediate future based on short-term historical data, our GRAINS proximity sensing system can effectively identify outliers brought by granular jamming.

**Localization and shape estimation of buried objects in GM**
In this section, we integrate GRAINS into a global exploration strategy based on BOA (text S7) to estimate the location and contour of buried objects in GM using proximity sensing. We envision that this strategy could provide an autonomous and safe way to detect subterranean objects for unmanned extraterrestrial robots in the future.

The simplified workflow of this BOA-guided pre-touch exploration strategy (BPES) is given in Fig. 1-F, and the detailed one can be seen in fig. S4. In general, BOA is responsible for suggesting the next target position to GRAINS after receiving observations in GM from GRAINS. At the same time, the BOA will predict the mean and variance distribution after each raking, showing the estimated shape of buried objects iteratively.

Therefore, after several probe explorations, the localization and shape estimation can be determined by BOA's estimated mean distribution, as shown in the first row of Fig. 5, where the white dashed lines outline the ground truth of the object. The higher value (e.g., dark red) indicates the area with a higher possibility of object existence, while the lower value (e.g., dark blue) indicates the region with a lower likelihood of the presence of objects. We observe that the subterranean objects are roughly outlined by BPES after 17 slides, where each slide is defined as one complete operation of GRAINS from the start position to the goal position as suggested by BOA. Following this process, the objects stay intact, with no breakages, benefiting from the proximity sensing capabilities of GRAINS. To substantiate the robustness of BPES, we conduct further experiments in various GMs (text S8) with buried objects being efficiently outlined in all trials. It should be noted that since GRAINS provides proximity sensing information, the estimated shape of the objects might appear larger than their actual physical dimensions.

**DISCUSSION**

Proximity sensing is important for unmanned construction robots to perceive and locate underground objects in granular materials. Although a variety of proximity sensors have been developed over the past decades, none of them can simultaneously detect the presence of objects beneath GM and estimate the location and shape of the buried objects before excavation. To this end, we propose a proximity sensing system for detecting the presence of objects in GM, namely GRAINS, as well as a BOA-based exploration strategy for locating and outlining the buried objects at the same time.

Our system employs Gaussian process regression to learn and analyze the force signals collected through the active raking of the probe along a spiral trajectory within a particulate matter. This process enables it to detect anomalies in the force readings that may signal the presence of a buried object. Consequently, we can autonomously perceive subterranean objects without visual and tactile feedback. Furthermore, our GRAINS exhibits remarkable robustness. It completes the parameter optimization of the search granules through the early offline parameter calibration. This not only reduces learning time but also enhances prediction accuracy and strengthens the robustness of proximity sensing. By integrating the GRAINS system into an exploration framework guided by BOA, we have successfully utilized proximity perception to localize and estimate the shape of buried objects.

This work provides a robust and autonomous solution that enables future uncrewed building robots to perceive objects beneath GM in extraterrestrial applications, as depicted in fig. S6. In addition, proximity sensing in GM also introduces a new strategy for scenarios requiring delicate handling of buried objects or the need to avoid direct contact, such as mine clearance or archaeological excavations. While our system is capable of operating in multiple granules — even in mixed matters — thanks to the offline parameter calibration process, we have yet to consider scenarios involving GM with varying packing densities. For instance, a pile of dry sand that has been watered can form highly compressed sand blocks. These situations pose challenges not only due to the physical properties of GM but also the complexity of the chemical bonds. Addressing these challenges may form part of our future work. Moreover, our current approach only utilizes force readings in the *x-y* plane, despite the presence of a 6-axis F/T sensor mounted on the robot arm. This presents two potential directions for future work: on one hand, we could design a low-cost 2D force gauge applicable to our current system. On the other hand, we could explore how to use the existing 6-axis F/T sensor to estimate the depth of buried objects using proximity sensing.

In summary, this work reports an autonomous proximity sensing system for detecting buried objects in granules. It empowers robots to sense the presence of subterranean entities without reliance on visual and tactile feedback, subsequently enabling them to estimate the location and shape of underground obstacles prior to excavation, all without human intervention.

## MATERIALS AND METHODS

### GRAINS prototype

As shown in Fig. 1-D, the prototype of GRAINS mainly includes three parts, i.e., the 6-DoF industrial robotic manipulator (UR5, Universal Robots, Denmark), the 6-axis F/T sensor (Force Torque Sensor FT 300, Robotiq, Canada), and a 3D-printed probe rod. The UR5 is controlled using Universal Robots ROS Driver on ROS kinetic, and the F/T sensor is driven by Robotiq meta-package. The probe is made of ABS (Acrylonitrile Butadiene Styrene) plastic with 1 cm diameter and 14 cm in length. All experiments are conducted via a computer with Intel$^©$ Core™ i7-6700 CPU and NVIDIA GTX 1080Ti GPU running Ubuntu 20.04.4 LTS.

### Granular materials

As illustrated in fig. S3-E, four types of granules are used in this investigation. One is golden sand composed of silica with diameters ranging from 0.5 to 1.6 mm. The second one is cassia seed in roughly rectangular sizes with around 3.5 mm in length and 2.5 mm in width. The third one is cat litter with polydisperse particles ranging in diameter from 3 to 4 mm. The last one is soybean with approximately 6.5 mm in diameter. The roughness of these four granular materials can be roughly arranged as: (rougher) sand > cat litter > cassia seed > soybean (smoother).

### Experimental methods

Methods employed in this study, e.g., Gaussian process regression, Bayesian optimization algorithm can be seen in Supplementary Materials for a complete description.

**Acknowledgments:**
   **Funding:**

   **Author contributions:**
   Z.Z. found phenomena, proposed ideas, designed and conducted experiments, analyzed data. R.J. developed robot control algorithms. Y.Y. provided critical feedback on the system and exploration strategy. R.H. S.J. provided critical feedback on learning-based methods. Q.J. provided critical feedback on mechanisms in granular materials. L.Z. provided full support for experiments. J.P. proposed the concept and led this project. Z.Z. completed the draft, and other authors reviewed and revised the manuscript.

   **Competing interests:** The authors declare that they have no competing interests.

   **Data and materials availability:** All data needed to evaluate the conclusions are available in the article and the Supplementary Materials.

## Figures

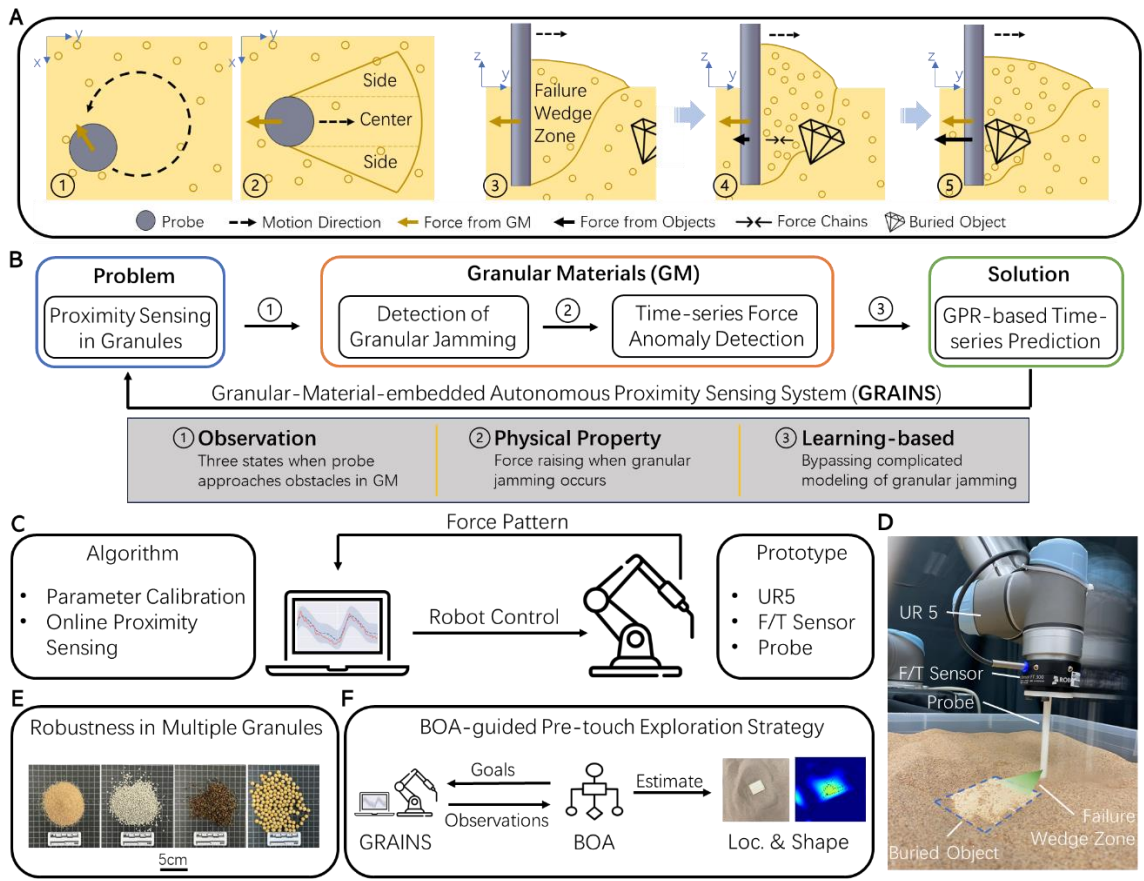

**Fig. 1. Overview of our work.** (**A**) Physical principles involved in granules. ① local granular fluidization by the circular vibration. ② linear plate drag. ① and ② constitute the proposed spiral trajectory for probe raking. ③, ④ and ⑤ illustrate three states when the probe linearly approaches buried objects in granules, that is, non-contact state, granular jamming state, and contact state, respectively. The length change of the arrows indicates the variations in force within *x-y* plane during the raking process. The concept of the failure wedge model is illustrated in ② and ③ from the top view and side view, respectively. (**B**) The proximity sensing problem in granules is converted to the real-time detection of the force anomaly caused by the granular jamming when objects enter the failure wedge zone in the vicinity of the probe based on the GPR time-series prediction. (**C**) Framework of the proposed GRAINS, including perception algorithm (see Fig. 3) and prototype (see (**D**)). (**E**) Four types of granules used for validating the robustness of GRAINS. (**F**) Bayesian-optimization-algorithm (BOA)-guided pre-touch exploration strategy (aided by GRAINS) for safely localizing buried objects and estimating their subterranean distribution.

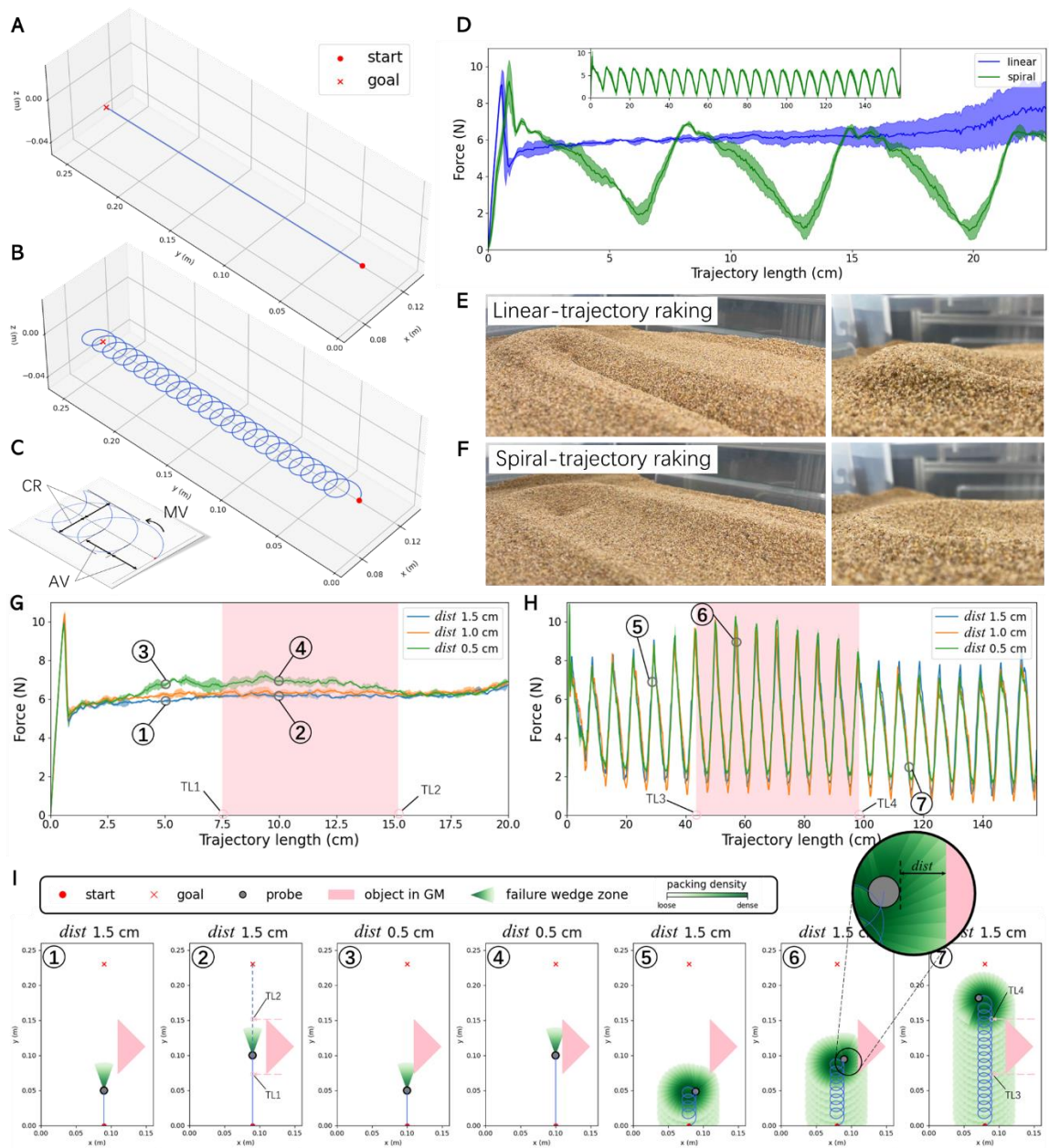

**Fig. 2. Raking trajectory.** (**A**) Linear trajectory. (**B**) Spiral trajectory, composed of linear advancement and circular vibration. (**C**) Parameters for defining a spiral trajectory: CR - circular radius, AV - advance velocity, and MV - motion velocity. (**D**) Drag force variations along linear (blue) or spiral (green) trajectories, when there are no subterranean objects buried in GM. The inset shows the drag force along the whole spiral trajectory and the shaded error bars indicate the ±1 SD. It is shown that around the goal position, the drag force along the spiral trajectory is more consistent than that along the linear trajectory. (**E**)(**F**) Left: Granular surfaces after linear and spiral raking, respectively; Right: zoomed view of the surfaces around the goal location after raking. (**G**)(**H**) Drag force variations along the linear and spiral trajectories, respectively, when a triangular brick is buried, as shown in (I). From (G), no response of the drag force is seen until *dist* = 0.5 cm. From (H), the response of drag force is evident when *dist* = 1.5 cm. Thus, a larger perception area for buried objects in GM when the probe rakes along the spiral trajectory. (**I**) Comparison of lateral perception areas of two types of trajectories. The differences in the perception areas can be qualitatively explained by the change of packing density using the failure wedge model. The zoomed view shows the definition of *dist*, which is the minimum distance between the buried object and the probe along the whole trajectory.

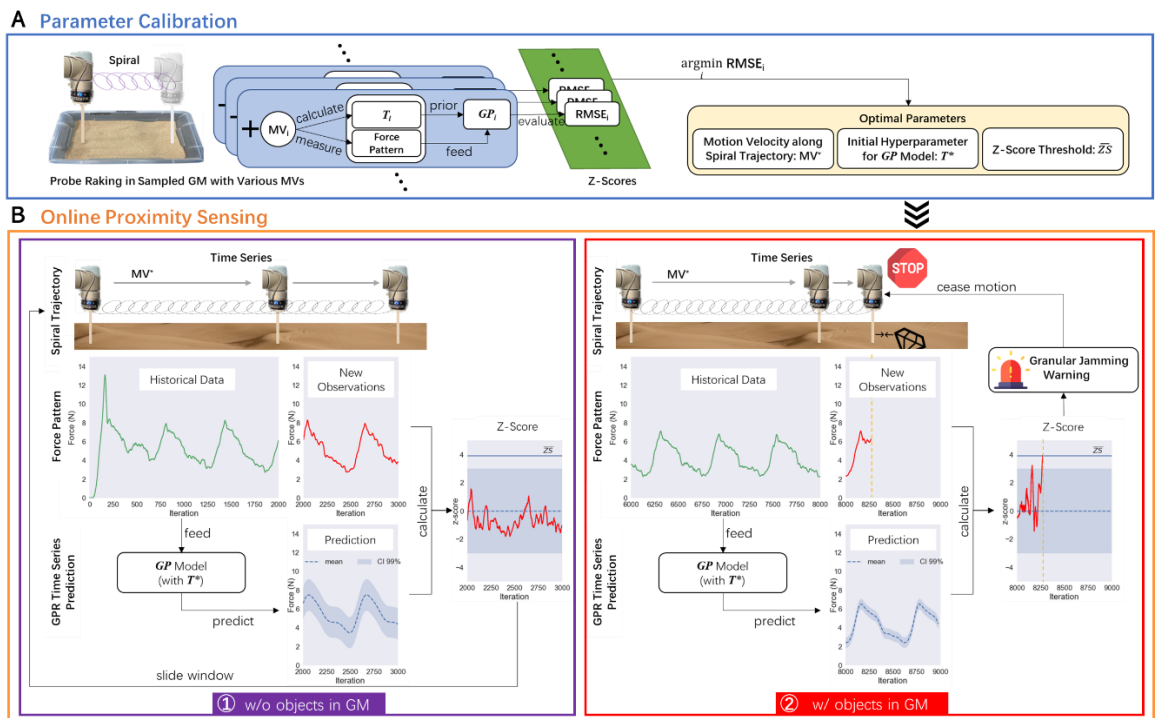

**Fig. 3. Perception algorithm in GRAINS.** (**A**) Parameter calibration. GRAINS automatically controls the probe with different MVs raking in GM sampled from granules that may contain buried objects. Then it evaluates the prediction performance of *GP* at each $MV_i$ by calculating the $RMSE_i$ of z-scores. Finally, after finding the minimum RMSE, GRAINS can determine corresponding optimal parameters. As revealed in text S4, these parameters would significantly affect performances of the next online proximity sensing. (**B**) Online proximity sensing. GRAINS detects the proximity of objects in GM through online detection of force anomalies caused by granular jamming, using Gaussian process regression. It learns the historical force patterns within a specific time window and compares the succeeding force measurements with the predicted values in real time. If no object is present in the proximity of the probe, the z-scores of new force readings should remain within a high confidence area (①). However, if the z-scores of certain force patterns diverge from the predicted range, GRAINS will issue a warning to the UR5 controller to stop the probe motion (②).

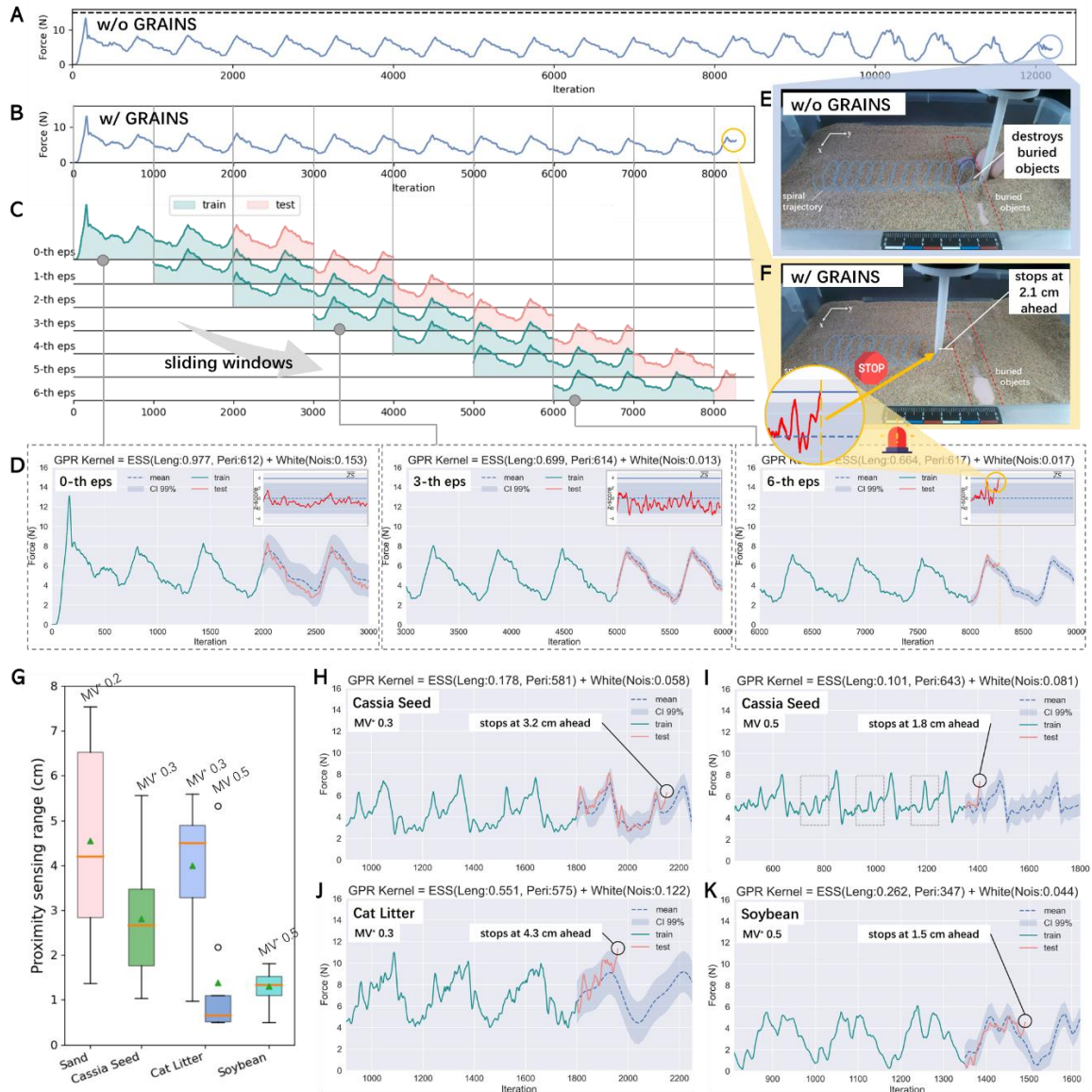

**Fig. 4. Proximity sensing experiments.** (**A**)(**B**) Force patterns measured in experiments with and without GRAINS, respectively. A fixed force threshold is given as the dash line in (A). (**C**) Online detection of proximity sensing by sliding windows in GRAINS. (**D**) The details of GPR learning and prediction results at 0, 3, 6-th episode (eps) of the experiment with GRAINS are given as well. Specifically, the force pattern in the first 2000 iterations is regarded as the training set and the predicted force pattern in terms of mean (dash line) and standard deviation (CI 99%) over the next 1000 iterations is given by the trained *GP* Model, whose optimized hyperparameters are also shown on the title of each subfigure. Since z-scores of the test set indicate the differences between real force data and predicted ones, once the z-score exceeds a threshold $\overline{ZS}$, GRAINS will stop the probe motion. (**E**) Snapshot when the probe destroys buried cylinders due to the absence of proximity sensing perception at the end of (A). (**F**) Snapshot when GRAINS detects the proximity sensing of ahead buried

objects at the end of (**B**). (**G**) Proximity sensing ranges in four types of granules. (**H**)-(**K**) Details of granular jamming identification in various GM.

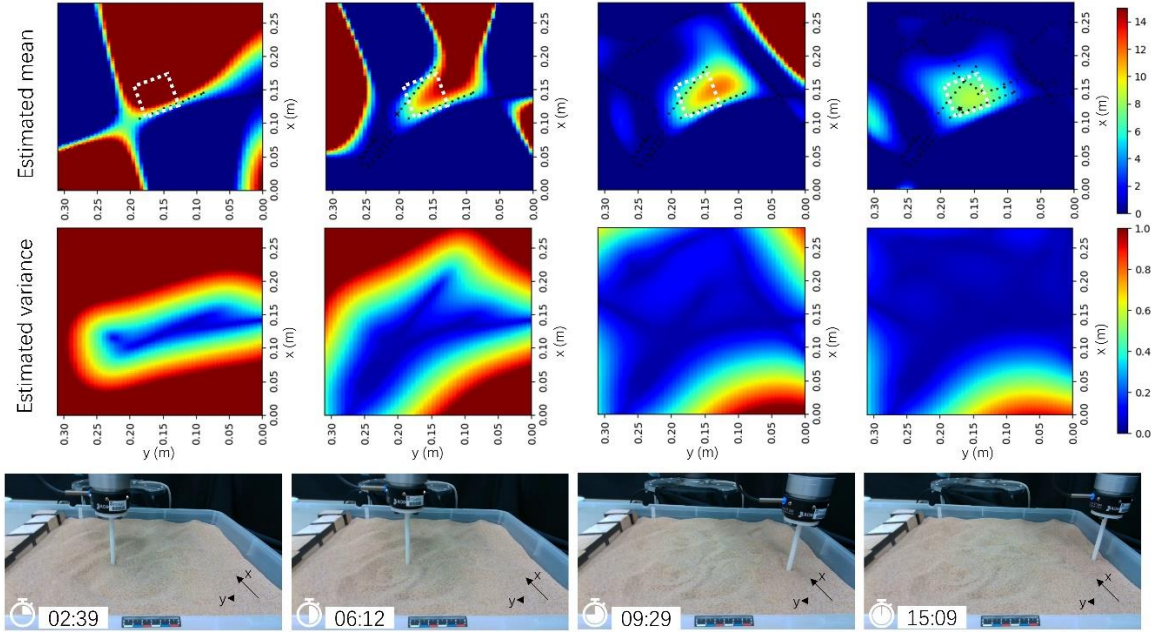

**Fig. 5. BOA-guided pre-touch exploration strategy using GRAINS.** A wooden square represented by white dashed lines in the first row is buried in the sand. After each exploration, the mean and variance are estimated by BOA accordingly. At the end of the experiment, the position of the buried square is localized, and its shape is approximately outlined by mean values. The real shape of buried object is outlined as dashed lines in the first row.

## Tables

**Table 1. A comparison between state-of-the-art proximity sensors used to detect buried objects in granular materials and our GRAINS.** NA: not applicable. NR: not reported. SLDV: scanning laser Doppler vibrometer. IMU: inertial measurement unit. F/T: force/torque sensor. A/S coupling: acoustic-to-seismic coupling. $^*$ In the strict sense, they belong to tactile sensors that need contact with buried stuff.

| Type | Sensor | Principle | Transduction Method | Sensing Range (cm) | Robustness in Multiple GM | GM properties considered | | | Feature | Level of autonomy for extraterrestrial scenario |
|---|---|---|---|---|---|---|---|---|---|---|
| | | | | | | Fluidization | Jamming | Failure Wedge | | |
| Non-invasive | Ground penetrating radar (21) | Electromagnetics | NA | Meter-level | NR | NA | NA | NA | Compact & robust, but vast post-processing work (22) and qualified operators required. | Medium |
| | EZiDIG (23) | Electromagnetics | NA | < 200 | NR | NA | NA | NA | | |
| | Frazier et al. (24) | Acoustics | Sonar array | < 100 | NR | NA | NA | NA | Autonomous, but complex (multi-sonar). | Low |
| | LAMBDIS (25) | A/S coupling | Laser, camera | NR | NR | NA | NA | NA | Accurate, but complicated and bulky. | |
| | Sugimoto et al. (26) | A/S coupling | Flat speakers, SLDV | < 10 | NR | NA | NA | NA | | |
| Invasive | Syrymova et al. (27) | Vibration | Vibro-tactile | 0* | ✓ | ✗ | ✓ | ✗ | Simple, but tactile information needed. | Low |
| | Digger Finger (28) | Vision | GelSight | 0* | NR | ✓ | ✗ | ✗ | | |
| | Jia et al. (29, 30) | Multi-modal | BioTac | 0* | ✓ | ✗ | ✓ | ✗ | | |
| | Jia et al. (31) | Force | F/T, IMU | NR | NR | ✗ | ✗ | ✓ | Simple, but many manual tuning parameters. | Low |
| | GRAINS (This work) | Force | F/T | 0.5 – 7 | ✓ | ✓ | ✓ | ✓ | Simple, autonomous, robust, and no contact. | High |

**Table 2. Parameter calibration.** GRAINS automatically rakes in multiple GM and selects the optimal $MV^*$ for each granule based on the minimum RMSE of z-scores (in bold). Then the corresponding maximum z-score (in bold) is considered the threshold for anomaly detection in the subsequent online phase. The periodicity prior $T_i$ at each $MV_i$ for the periodic kernel in *GP* Model is also calculated.

| GM | $MV_1(0.2) / T_1(439)$ | | $MV_2(0.3) / T_2(293)$ | | $MV_3(0.4) / T_3(220)$ | | $MV_4(0.5) / T_4(176)$ | | $MV_5(0.6) / T_5(146)$ | | $MV_6(0.7) / T_6(125)$ | |
|---|---|---|---|---|---|---|---|---|---|---|---|---|
| | RMSE | max | RMSE | max | RMSE | max | RMSE | max | RMSE | max | RMSE | max |
| Sand | **0.9191** | **3.9** | 1.3739 | 7.0 | 1.4695 | 8.7 | 1.2284 | 4.2 | 1.9801 | 12.4 | 1.2498 | 4.3 |
| Cassia Seed | 1.4971 | 8.4 | **1.0110** | **5.0** | 1.2832 | 4.9 | 1.1968 | 5.0 | 2.0234 | 8.9 | 1.0395 | 3.3 |
| Cat Litter | 1.0992 | 4.7 | **1.0741** | **5.0** | 1.2340 | 7.4 | 1.2565 | 4.8 | 1.3190 | 5.9 | 1.1983 | 5.1 |
| Soybean | 1.1591 | 5.8 | 1.2548 | 3.6 | 1.3345 | 5.6 | **1.0488** | **4.8** | 1.1644 | 3.5 | 1.3709 | 4.4 |

**Supplementary Materials**

**Text S**1. Typical Proximity Sensors Working in Air and Water
**Text S**2. Comparison to State-of-the-art Sensors in GM
**Text S**3. Effect of Packing Density on Jamming Force Detection
**Text S**4. Effect of Spiral Trajectory on Proximity Sensing
**Text S**5. Gaussian Process Regression
**Text S**6. Z-scores and Confidence Interval
**Text S**7. Bayesian Optimization Algorithm
**Text S**8. BOA-guided Pre-touch Exploration Strategy
**Fig. S**1. Qualitative study about the effect of packing density on jamming force detection
**Fig. S**2. Effect of spiral trajectory on proximity sensing
**Fig. S**3. Force readings in multiple granules with different particle sizes and roughness
**Fig. S**4. Workflow of the BOA-guided pre-touch exploration strategy using GRAINS
**Fig. S**5. BOA-guided pre-touch exploration strategy
**Fig. S**6. A conceptual diagram of unmanned construction robots with GRAINS to detect objects on the granular crust of Mars
**Table S**1. Representative proximity sensors operating in air and water
**Table S**2. Measured periodicity and calculated periodicity prior with CR 0.02 m and AV 0.01 m, but at different MVs
**Movie S1**. Overview of this work
**Movie S2**. Physical principle
**Movie S3**. Effect of packing density on jamming force detection
**Movie S4**. Raking trajectory
**Movie S5**. Framework of GRAINS
**Movie S6**. Proximity sensing experiments
**Movie S7**. Localization and shape estimation of buried objects in GM

**Text S1. Typical Proximity Sensors Working in Air and Water**

Currently, numerous studies employ proximity sensors for robotic tasks in the air, such as grasping, but a few of them are designed for underwater scenarios (see table S1). According to different principles and operating modes, we try to introduce several representative proximity sensors and hope to give readers a general understanding of them.

Proximity sensors based on vision (*45*), infrared (*46*), ultrasound (*47*), and time-of-flight (*48*) are widely used in daily-lift applications, where vision-based one belongs to passive mode yet others are all in active mode since they need to transmit and receive some kinds of signals. In addition, the triboelectric effect has also been utilized to respond to touchless stimulation (*49*). Some study investigates capacitance to sense human gesture in the air (*50*).

Due to high distortion and poor light underwater, a few proximity sensors are reported in liquid environments. Depending on marine conditions, some sensors based on vision (*51*) and acoustics (*52*) could provide near-field perception over relatively long distances. Furthermore, according to the amount of power, the ship and submarine could employ sonar (sound navigation and ranging) to perceive environments under the surface of the water to assist navigation. As for underwater proximity sensing, humans mainly learn from the sensory organs of marine creatures. For example, after the observation of flow-sensing from wavy seal whiskers, authors in (*53, 54*) design a 3D-printed whisker to feel the flow and upstream wake nearby. Similarly, inspired by the lateral line system of fish, several investigations propose elaborated proximity sensors on account of the pressure gradient (*55, 56*). Additionally, some studies have introduced SensorPod (*35*) and Slender Probes (*57*) that use electrolocation to achieve near-field sensing, taking cues from weakly electric fish in the dark deep oceans (*58*).

**Table S1. Representative proximity sensors operating in air and water.** *from Murata Inc. †from STMicroelectronics Inc. ‡from TE Connectivity Inc. (IRED: infra ted emitting diode. TENG: triboelectric nanogenerator. DVL: Doppler velocity log. LCP: liquid crystal polymer.)

| Working Medium | Sensor | Principle | Sensing Range (cm) | Operating Mode | Transduction Method |
|---|---|---|---|---|---|
| Air | ProTac (*46*) | vision | $2 - 10$ | passive | fish-eye camera |
| | Sensor Skin (*47*) | infrared | $0 - 25$ | active | IRED+pin diodes |
| | UltraHaptics (*48*) | ultrasound | $20 - 40$ | active | 320 MA40S4S units* |
| | Spherical Fingertip (*49*) | time-of-flight | $1 - 15$ | active | 5 VL6180X sensors† |
| | FBSS (*50*) | triboelectric effect | $0 - 2$ | passive | TENG |
| | Gesture-Sensing (*51*) | capacitance | $1 - 15$ | active | receiver & transmitting electrodes |
| Water | CUREE (*52*) | vision | $< 150$ | passive | stereo camera |
| | HAUV (*53*) | acoustics | $\sim 100$ | active | DVL+sonar |
| | Zheng *et al.* (*55*) | vibration | $2 - 8$ | passive | 3D-printed whisker |
| | ALLS (*56*) | pressure | $\sim 5$ | passive | 6 MS5407-AM sensors‡ |
| | Kottapalli *et al.* (*57*) | pressure | $5 - 20$ | passive | LCP array |
| | SensorPod (*35*) | electrolocation | $10 - 25$ | active | excitation & sensing electrodes |
| | Slender Probes (*58*) | electrolocation | $0 - 22$ | active | emission & perception electrodes |

**Text S2. Comparison to State-of-the-art Sensors in GM**

Table 1 exhibits a comparison between state-of-the-art proximity sensors and GRAINS.

As a non-invasive method, ground penetrating radar (*21*) is a common tool in construction and geophysical applications. It transmits electromagnetic radiation and detects underground stuff by interpreting reflected waves. Even if it could be a compact system and provide robust results at meter level, it suffers from massive post-processing work (*22*) and requires qualified people to operate it. Based on the same principle, some cable avoidance tools (CAT), like EZiDIG (*23*), can be mounted on the excavator to increase excavation confidence. However, it only responds to metallic objects in the soil such as underground water pipes and cables, and also requires operators to be trained in advance. The use-cost of these devices increases as a result. In addition, based on acoustics, (*24*) provides a sensing system using a set of complicated sonar arrays toward the ground. Due to different principles, aforementioned sensors allow meter-level perception, but a group of laser-acoustic sensors, e.g., (*25, 26*), only feel the presence of buried utilities within several centimeters. Normally, these systems are complicated and bulky, since they mainly contain acoustic sources, the laser generator, the high-power supply, and other accessory equipment.

In comparison, invasive sensing systems are quite simple and lightweight. And they mainly employ some GM properties to perceive unseen objects, e.g., granular fluidization, jamming, and failure wedge zone, as in Table 1. For instance, several studies primarily focus on the haptic modality. Syrymova et al. (*27*) introduce a vibro-tactile method that classifies the absence or presence of a rigid body in GM based on mechanical vibrations by squeezing granules in a rubber balloon. Utilizing the vision-based tactile sensor, GelSight (*59*), the Digger Finger in (*28*) estimates the shape of simple objects in GM by penetrating into GM and acquiring touch feedback. Jia et al. (*29, 30*) use multiple tactile modalities from the haptic sensor BioTac (SynTouch LLC, Los Angeles, CA) to estimate contact states between robotic fingertips and objects in GM. Obviously, these tactile techniques require direct contact between the sensory surface and buried objects, and can not sense obstacles in advance. Most related work on proximity sensing in GM comes from (*31*), which is based on the characteristic of force variation between tool and object and employs a pre-defined failure wedge model to automatically estimate the distribution of underground stuff. It can be found this work is highly constrained to a specific GM since lots of manual tuning parameters are used in the sensor model based on preliminary experiments and empirical equations. For other granules, those parameters may not work anymore.

Inspired by (*28, 31*), we propose the GRAINS with a force gauge and a probe rod. It takes advantage of automatic offline parameter calibration to achieve robust proximity sensing about 0.5 ~ 7 cm in multiple particle matters. Then, it has shown that the Gaussian process regression could flexibly learn the force characteristics from GM and automatically detect the granular jamming state by the force anomaly resulting from buried objects. In this work, we observe and investigate granule properties, i.e., granular fluidization, granular jamming, and failure wedge zone, instructing us to propose the spiral raking trajectory, periodic force pattern, etc.

For extraterrestrial applications, due to the lack of atmosphere, these acoustic-related devices (*24-26*) do not work anymore. Furthermore, tactile modality is not beneficial for safe operation facing unseen buried obstacles (*27-30*). Many manual tuning parameters in (*31*) also highly constrain its applicability in automatic extraterrestrial scenarios. Benefiting from the high robustness and compact size, the ground penetrating radar and its variation have been mounted on existing teleoperated rovers (e.g., Yutu-2 (*60*)) to complete the extraterrestrial exploration. Currently, human-in-loop control is still needed for these vehicles to finish tasks. However, in the future, the construction of exoplanet bases needs intelligent autonomous robots, and our GRAINS offers a simple yet reliable system to sense underground obstacles without human intervention.

**Text S3. Effect of Packing Density on Jamming Force Detection**

Granular jamming occurs as the packing density (p.k.d.) of particles increases (i.e., more particles per unit volume) (*8*). Therefore, we expect that a higher packing density will result in the earlier detection of the jamming force. To test this, we drag a probe through GM with high and low packing densities and record the force readings until the rod hits the same object. As a comparison, we demonstrate two extreme cases where one is a rigid body and the other has nothing but air between the probe and the obstacle to show the force delivery (see fig. S1-A).

As revealed in fig. S1-B, a higher packing density of GM results in the probe detecting the buried object earlier, which in our experiment means when the measured force exceeds the stop bar (8N). It is shown that when there is nothing but air between the probe and the obstacle, the probe hits it after 5s, and the growth gradient is extremely large, as presented in the blue curve. On the other hand, if a solid object is between them, the force on the probe (red curve) quickly increases and exceeds the stop bar before 2s with a large raising gradient similar to that in the case of air. Furthermore, we test the same motion among dry sand, whose packing density $\phi$ is determined by the height of granular particles in the container as

$$\phi = M / (\rho A h), \quad (S1)$$

where $M$ and $\rho$ are the mass and density of sand and $A$ is the bottom area of the container. $h$ refers to the height of sand occupied in the container and could be measured from the side of view, as shown in fig. S1-A. In this test, $h$ in the case of low p.k.d. and high p.k.d. are roughly 4.5 cm and 3.5 cm, respectively. From fig. S1-B, we can find that the drag force in high p.k.d. GM (green curve) increases earlier and also hits the stop bar earlier than that in low p.k.d. GM (orange curve). From this qualitative study, it can be concluded that a higher packing density of GM between the probe and the buried object results in the force from the buried object being transmitted to the probe more quickly. It can be explained by the granular jamming process, where it requires a longer distance in loose GM than that in dense GM to squeeze particles to solidify themselves and further generate force chains among granules to deliver forces from obstacles. Therefore, to help the probe more accurately perceive objects nearby, the high packing density in the vicinity of the probe is desired. Thus, the raking trajectory for the probe that could generate high p.k.d. around it should be well studied.

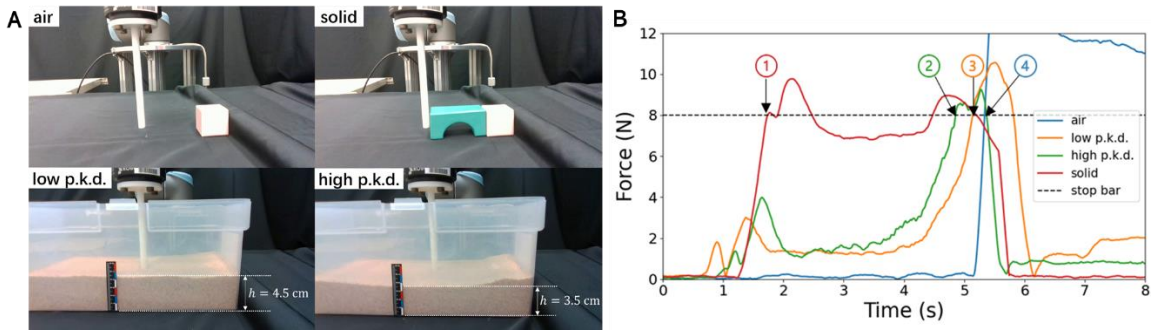

**Fig. S1. Qualitative study about the effect of packing density on jamming force detection.** (**A**) Experiment setup. (**B**) A higher packing density will result in the earlier detection of jamming force.

**Text S4. Effect of Spiral Trajectory on Proximity Sensing**

From Fig. 2-C, we can see that the parameters CR, AV and MV uniquely determine the spiral trajectory, where CR and AV determine the shape of the spiral path, while MV manipulates the timing schedule. In the following part, we will investigate how these parameters affect proximity sensing in granules.

*Sensing area*

In this subsection, we will show that sensing areas vary with different CR and AV values. In addition, we do not observe a clear relationship between the proximity sensing range and the different (CR, AV) sets.

In fig. S2-A, we demonstrate 9 spiral paths determined by different groups of CR and AV parameters displayed in each figure title. It can be observed that CR is responsible for the sensing width and AV decides the aggression of probe advance. Specifically, with the growth of CR, the sensing width increases. And the larger AV is, the more aggressively the probe explores forward. In addition, after our investigation of the proximity sensing range along these spiral paths, we do not find a direct relationship between the sensing range and the set of CR and AV. In other words, if we want to search through GM conservatively, we should choose small CR and AV values. But if we would like to explore granules quickly, then large values of CR and AV are desired. In both scenarios, GRAINS would provide similar proximity sensing results.

*Periodicity prior*

In addition, the MV, a dimensionless parameter ranging from 0 to 1, will determine the motion velocity of the probe moving along the spiral path given by CR and AV. However, MV could dramatically affect the performance of GRAINS. On the one hand, MV determines the periodicity prior used in the periodic kernel of *GP* Model. A good prior is beneficial for reducing the computational costs in GPR. On the other hand, MV has non-negligible influences on the force data collection and then affects sensing accuracy. In this subsection, we first introduce the effects of MV on periodicity prior, followed by its influences on sensing accuracy in the next subsection.

As reported in (*61*), there exists stick-slip phenomenon in GM, where GM exhibits purely periodic fluctuations when a probe rakes in GM with relatively shallow depth. As shown in Fig. 2-D, similar periodic force variation can also be found in our experiments. However, after investigations, we find that this is not due to the stick-slip mechanism, but instead results from the circular motion involved in spiral trajectories. Furthermore, we figure out that the periodicity in our force data is related to the MV. Specifically, the period in force patterns is the time (more accurately, the measurement iterations from the F/T sensor) that the probe takes to complete a circular motion, i.e., one cycle, on the spiral paths. Three cycles of spiral paths with different CR and AV values are shown in fig. S2-A. Given one spiral path, different MVs lead to different durations to complete a cycle, thus leading to the different periodicity of force patterns. In GRAINS, we hope to learn force patterns through the *GP* Model, among which the periodic kernel is responsible for learning the periodicity of force patterns. If we can acquire the approximate period before training, it will greatly reduce the training time and improve the training effect of *GP*.

Fortunately, we can calculate this periodicity prior $T$ before training with the following equation:

$$T = \frac{\|\text{path}(CR, AV)\|}{V_0 * MV} f_s, \quad MV \in [0,1], \tag{S2}$$

where the numerator refers to the length of one cycle in the spiral path determined by CR and AV, the denominator is the speed of probe motion, and $f_s$ is the sampling frequency of the F/T sensor. Here $V_0 = 0.08968$ and $f_s = 62.5$ Hz in GRAINS. Table S2 exhibits measured periodicity from force data and periodicity prior calculated by the above equation, and we can find the calculated periodicity priors can well estimate the real measured periodicity.

Based on the periodicity prior, the *GP* Model provides a better prediction result with low computational costs, as demonstrated in fig. S2. Here the spiral trajectory is defined by CR 0.02 m, AV 0.01 m, and MV 0.2. The fig. S2-B and C demonstrate GPR predictions in the first three episodes with and without periodicity prior, respectively. It is obvious that the prediction results from the *GP* Model based on the periodicity prior outperform that without using periodicity prior. The time costs of GPR in each episode are given as fig. S2-D as well. It can be found that the periodicity prior helps the *GP* Model to quickly learn force patterns from historical data and provide predictions very soon.

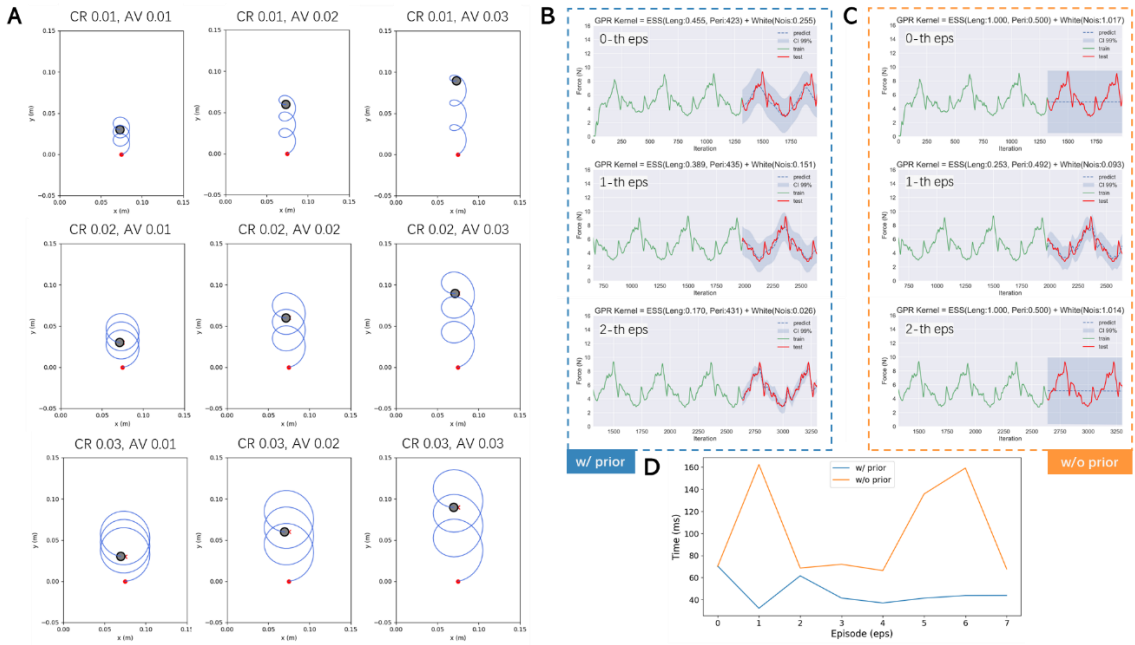

**Fig. S2. Effect of spiral trajectory on proximity sensing.** (**A**) Spiral trajectories with different sets of CR (unit: m) and AV (unit: m). (**B**)(**C**) Predictions of three episodes from GPR with and without periodicity priors, respectively, based on the same force patterns. (**D**) Computational costs of GPR about the learning and prediction with and without periodicity priors. It is worth noting that learning and prediction is the most time-consuming part of GRAINS. Here we can see the average time cost of GPR's learning and prediction with periodicity priors is about 50 ms (i.e., ~ 20 Hz), allowing GRAINS works in a real-time manner.

**Table S2. Measured periodicity and periodicity prior with CR 0.02 m and AV 0.01 m, but at different MVs.**

| MV | 0.2 | 0.3 | 0.4 | 0.5 | 0.6 | 0.7 |
|---|---|---|---|---|---|---|
| Measured Periodicity | 435 | 289 | 216 | 173 | 144 | 124 |
| Periodicity Prior | 439 | 293 | 220 | 176 | 146 | 125 |

*Sensing accuracy*

When investigating GRAINS raking in different GMs, we observe significant differences in GPR's computational cost and prediction performance. These differences ultimately affect the accuracy of proximity sensing and buried object detection. Through experiments, we discover that the variance originates from motion velocity MV, a dimensionless variable ranging from 0 to 1. MV controls the UR5 end-effector's speed along the path determined by CR and AV. To further clarify this phenomenon, here we conduct two additional experiments that use the same sampling frequency of the F/T sensor.

The first experiment shows that random vibrations in force measurement are related to granule sizes (fig. S3-E). Four different GMs with varying particle sizes were tested, and it is known from RFT (*17, 18*) that the drag force acting on the probe is influenced by random contacts between granules and probe. Granules with numerous particle sizes exhibit different randomness in drag forces from the probe, with tiny particles such as sand densely occupying the area around the probe, resulting in continuous contact and smooth force readings (fig. S3-A). However, large-size particles such as soybean sparsely distributed around the probe with relatively big gaps, causing discontinuous collisions and high random vibrations in force measurements (fig. S3-D). Therefore, if force data with many random errors, such as in soybean, are fed into the GPR, it may overfit with a high computational cost, as the *GP* Model may be confused by many useless details from random errors. This can lead to increased false positives and disastrous detection accuracy for GRAINS.

To improve the detection accuracy of objects buried in large-size granules, we conduct the second experiment investigating the relationship between random vibrations and MV. We deploy GRAINS in a bed of soybeans with MVs ranging from 0.2 to 0.7. By comparing fig. S3-D and F, it can be observed that the higher MV led to smoother force measurements. This occurs because, with the same measurement frequency, higher MV is equivalent to downsampling data at low MV, which can reduce random errors accordingly.

Based on these two experiments, we can claim that different GMs display varying random errors in force measurements. Increasing MV is beneficial for reducing these errors, which aids in the learning and prediction of GPR. However, setting a high MV in GRAINS is not feasible due to the increased risk of collision between the probe and buried objects. Therefore, it is necessary to determine an appropriate MV for each GM. This can be achieved through offline parameter calibration to obtain data and find the optimal MV for each GM to be used in subsequent online phases. The calibration operation will significantly improve the detection accuracy of GRAINS in various GMs.

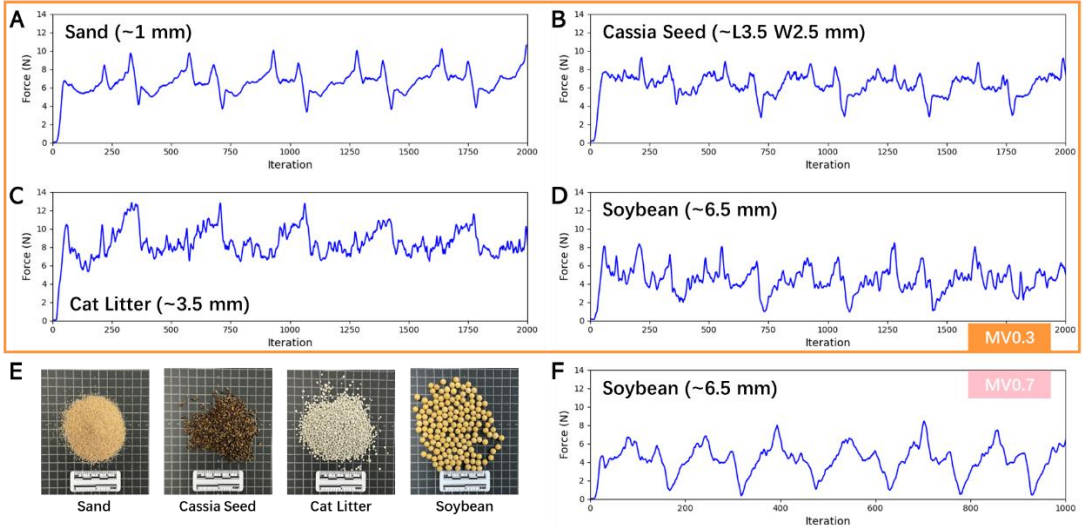

**Fig. S3. Force readings in multiple granules with different particle sizes and roughness.** (**A**)(**B**)(**C**)(**D**) exhibit drag force sequences along an identical spiral trajectory (CR: 0.03 m, AV: 0.03 m, MV: 0.3) for different granular materials in (**E**), including sand, cassia seed, cat litter, and soybean. We can observe that random errors increase with the growth of particle size. (**F**) Force measurements along the same path as (D) but with MV = 0.7. We can observe that random errors decrease when MV increases.

**Text S5. Gaussian Process Regression**

In this paper, we utilize Gaussian Process Regression (GPR) to model the soil-tool interaction. We provide a brief introduction to GPR here, and more details can be found in Chapter 2 of (*39*).

A Gaussian process ($GP$) is a collection of random variables, with any finite number of them having a joint Gaussian distribution. Mathematically, if we define the mean function $m(\boldsymbol{t})$ and covariance function $k(\boldsymbol{t}, \boldsymbol{t}')$ of a real process $p(\boldsymbol{t})$, then we can represent the $GP$ as:

$$p(\boldsymbol{t}) \sim GP(m(\boldsymbol{t}), k(\boldsymbol{t}, \boldsymbol{t}')) \tag{S3}$$

For the sake of notation simplicity, we will assume that the mean function is zero, i.e., $m(\boldsymbol{t}) = 0$.

The testing dataset $\mathcal{D} = \{(t_i, f_i) | i = 1, \cdots, n\}$ consists of $t_i \in \mathbb{R}$ as time series and $f_i \in \mathbb{R}$ as the force magnitude measured from the force sensor. The force value $f_i$ is related to every $t_i$ as follows:

$$f_i = p(t_i) + \epsilon_i, \ \forall i = 1, \cdots, n. \tag{S4}$$

Here, we assume that $\epsilon_i$ is independently and identically distributed Gaussian noise, drawn from a zero mean Gaussian distribution with variance $\sigma_n^2$, i.e., $\epsilon_i \sim \mathcal{N}(0, \sigma_n^2)$.

For a given set of test inputs $\boldsymbol{t}^*$, the posterior predictive distribution $p(\boldsymbol{t}^*)$ conditioned on $\mathcal{D}$ is a Gaussian distribution. The GPR can be expressed as follows:

$$\begin{aligned}
\boldsymbol{p}_* | \mathcal{D} &\sim GP(\bar{\boldsymbol{p}}_*, \text{cov}(\mathbf{p}_*)), \\
\bar{\boldsymbol{p}}_* &= K(\boldsymbol{t}_*, \boldsymbol{t})[K(\boldsymbol{t}, \boldsymbol{t}) + \sigma_n^2 I]^{-1} \boldsymbol{f}, \\
\text{cov}(\mathbf{p}_*) &= K(\boldsymbol{t}_*, \boldsymbol{t}_*) - K(\boldsymbol{t}_*, \boldsymbol{t})[K(\boldsymbol{t}, \boldsymbol{t}) + \sigma_n^2 I]^{-1} K(\boldsymbol{t}, \boldsymbol{t}_*).
\end{aligned} \tag{S5}$$

Here, $n$ denotes the number of training data and $n_*$ denotes the number of test data. The matrix $K(\boldsymbol{t}, \boldsymbol{t}_*)$ is an $n \times n_*$ matrix containing covariances of each pair of training and test points. Similarly, $K(\boldsymbol{t},\boldsymbol{t})$, $K(\boldsymbol{t}_*,\boldsymbol{t})$, and $K(\boldsymbol{t}_*,\boldsymbol{t}_*)$ are matrices of covariances between training points, between test points, and within test points, respectively.

In this work, the probe conducts circular motion in addition to linear raking. As a result, the measured forces exhibit a periodic pattern in the time domain, which can be modeled using the periodic kernel:

$$k_{per}(t_i, t_j) = \sigma^2 \exp\left(-\frac{2}{l^2}\sin^2\left(\pi\frac{|t_i - t_j|}{T}\right)\right). \tag{S6}$$

Here, $\sigma^2$ is the overall variance, $l$ is the length scale, and $T$ refers to the periodicity of the kernel. To account for the high randomness from granular particles, the white kernel is used to estimate the noise of the measured forces:

$$k_{whi}(t_i, t_j) = \sigma^2 I, \tag{S7}$$

where $\sigma^2$ is the noise variance and $I$ is the identity matrix. The sum-kernel, $k = k_{per} + k_{whi}$, is used as the prior for the covariance function in Eq. (S3). Additional kernel descriptions can be found in Chapter 2 of (*62*). For GPR training and prediction, the Python package *Scikit-Learn* (*63*) is used in this work.

**Text S6. Z-scores and Confidence Interval**

The z-scores, also known as standard scores in statistics, require knowledge of the mean and standard deviation of the complete population to which a data point belongs. In the GPR model, the predicted force values are given in terms of mean and standard deviation. Therefore, when a new force reading $x_i$ is obtained, its z-score can be calculated as follows:

$$z_i = \frac{x_i - \mu_i}{\sigma_i}, \tag{S8}$$

where $\mu_i$ and $\sigma_i$ are the predicted mean and standard deviation at the $i$-th iteration. The z-score indicates the deviation distance of the raw data $x_i$ from the population mean $\mu_i$, in units of the standard deviation $\sigma_i$.

In practice, for the normal distribution, the z-scores for 95% and 99% confidence interval (CI) are 1.96 and 2.576, respectively. Here, CI represents the probability that a parameter falls within a certain range of values. For example, if the z-score of a sampled data is 3, larger than $z_{CI=99\%} = 2.576$, it means this data is very unlikely to occur (with a chance less than 1%). In other words, this data may not belong to the given class with a high degree of certainty. Therefore, z-scores can be used to detect anomalies.

**Text S7. Bayesian Optimization Algorithm**

Proximity sensing in granular media can only provide local and sparse information about buried objects. Therefore, a smart and efficient exploration policy is required. In this paper, we use the Bayesian Optimization Algorithm (BOA) as the explorer to quickly locate and outline buried objects.

BOA is a powerful global optimization method for black-box functions (*64*) that are expensive to query and/or cannot be formulated in closed form. It consists of two components: a surrogate model is used to estimate the objective function to be optimized, and an acquisition function is used to identify the next best point for exploration.

In this study, the stiffness distribution in granular materials is treated as an unknown function, where the presence of buried objects refers to a high stiffness value and the absence of objects refers to a low stiffness value. In this experiment, we describe the presence information as $7$ and the absence information as $0$, as revealed in test S8 as well. Then the BOA is used to find the maximum stiffness corresponding to buried objects. The *GP* is used to provide a prior for this unknown stiffness distribution, and BOA utilizes the mean $\mu(\mathbf{x})$ and variance $\sigma^2(\mathbf{x})$ of the predictive *GP* posterior to identify the most likely location of the global maximum using an acquisition function. We use the squared-exponential (SE) kernel (*62*) in *GP* and the expectation improvement (EI) (*65*) as the acquisition function. The SE kernel is defined as

$$k_{SE}(\mathbf{x}_i, \mathbf{x}_j) = \sigma_{SE}^2 \exp\left(\frac{-\|\mathbf{x}_i - \mathbf{x}_j\|^2}{2l^2}\right), \tag{S9}$$

where $x_i, x_j \in \mathbb{R}^2$ represent 2D coordinates, $\sigma_{SE}^2$ is the kernel variance, and length scale $l$ represents the kernel width. The EI acquisition function can be expressed analytically as

$$EI(\mathbf{x}) = \begin{cases} (\mu(\mathbf{x}) - y^+)\Phi(\mathcal{Z}) + \sigma(\mathbf{x})\phi(\mathcal{Z}) & \text{if } \sigma(\mathbf{x}) > 0 \\ 0 & \text{if } \sigma(\mathbf{x}) = 0, \end{cases} \tag{S10}$$

where $\mathcal{Z} = (\mu(\mathbf{x}) - y^+)/\sigma(\mathbf{x})$, $y^+$ is the current maximum. Then $\phi(\cdot)$ and $\Phi(\cdot)$ refer to the probability density function (PDF) and the cumulative density function (CDF) of the standard normal distribution, respectively. We implement BOA to guide the probe motion using a Python package in (*66*).

**Text S8. BOA-guided Pre-touch Exploration Strategy**

Fig. S4 represents the workflow of BOA-guided pre-touch exploration strategy (BPES) using GRAINS in detail. We assume that there were no underground objects at the initial position $\mathbf{x}_{init}$, allowing the probe to safely penetrate the GM at the beginning. Initially, the BOA assigns a random goal to GRAINS, denoted by $\mathbf{x}_g$, since it has no prior information. As the probe advances, GRAINS will report the observations in GM. Specifically, it sends the absence information (defined by 0) to BOA at every 1 cm position during raking if no granular jamming is detected. If the probe approaches the goal $\mathbf{x}_g$ without detecting the granular jamming, the acquisition function in BOA will generate a new target with the highest probability of object existence at the current stage and update it to GRAINS.

If granular jamming is detected during the plowing, GRAINS will report the presence signal (defined by 7) to BOA and stop the probe motion accordingly. The goal and start positions would be exchanged, and the new start position (i.e., previous goal position) is assigned as a penetration candidate position $\mathbf{x}_e$, then the current point $\mathbf{x}_c$ is recorded as a proximity sensing point $\mathbf{x}_p$. After that, the probe will leave GM and move the start point overhead, then start raking through GM again. It should be noted that before raking, there is a penetration process at $\mathbf{x}_e$. If the probe successfully penetrates GM at $\mathbf{x}_e$, GRAINS will control the probe to the goal. As anticipated, the probe would detect the granular jamming again at a particular position, indicating the other boundary of the object we have just found, or another object located on the exploration path. Then penetration candidate position $\mathbf{x}_e$ would be updated to the point that one step from the current position to $\mathbf{x}_g$. If the probe can not invade the GM at $\mathbf{x}_e$, then $\mathbf{x}_e$ will also move forward one step to the goal point $\mathbf{x}_g$ until $\mathbf{x}_e$ gets close within a range of the proximity sensing point $\mathbf{x}_p$. If so, the probe will return to the initial position $\mathbf{x}_{init}$ or the first penetration point $\mathbf{x}_e^1$.

*Experiment in sand*
In this experiment, we test BPES in a bed of dry, loose sand, as represented in fig. S5-A, where a single square wooden brick (see fig. S5-C) is buried inside. Here the sandbox is around $49 \times 47$ cm and we set a search area (outlined in dashed green lines) whose size is smaller than that of the sandbox in order to eliminate the boundary effect with the container wall (*67*). Note that the buried object is located in this search area. The experiment process is shown in the supplementary video (movie S7) and the resulting estimated means and variance at 4 moments are reported in Fig. 5.

*Experiment in cassia seed*
In this experiment, we test BPES in a bed of cassia seeds, as illustrated in fig. S5-B, where two different objects are buried in the smaller search area, as depicted in fig. S5-D and E, respectively.

From fig. S5-D, it is shown that BPES successfully identifies two objects in cassia seeds after 15 raking slides, whose shapes are relatively larger than the ground truth (revealed in black lines). Nevertheless, fig. S5-E shows resulting distributions and manifests that two objects are not distinguished even after 20 raking slides, probably because they are quite close to each other, and the probe may not recognize the gap between them during the exploration. Even so, the overall shape of these two objects is still roughly estimated.

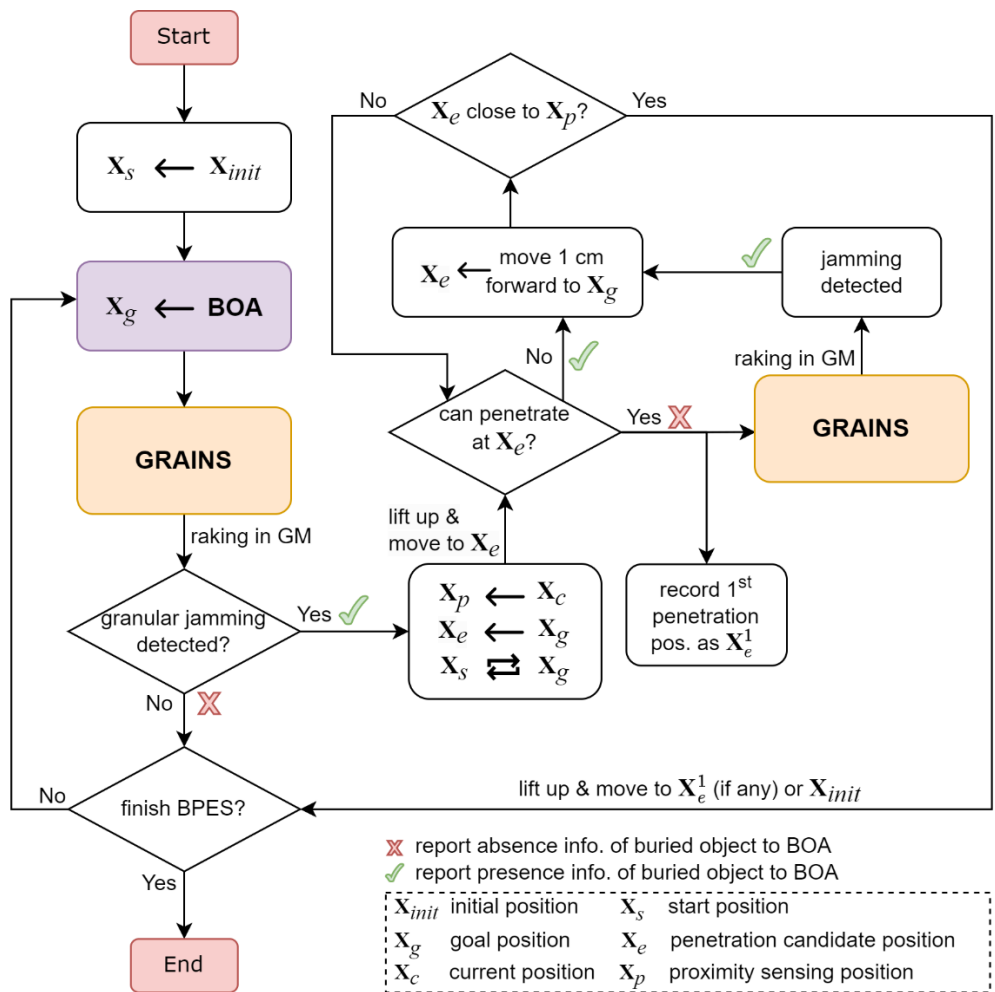

**Fig. S4. Workflow of the BOA-guided pre-touch exploration strategy using GRAINS.**

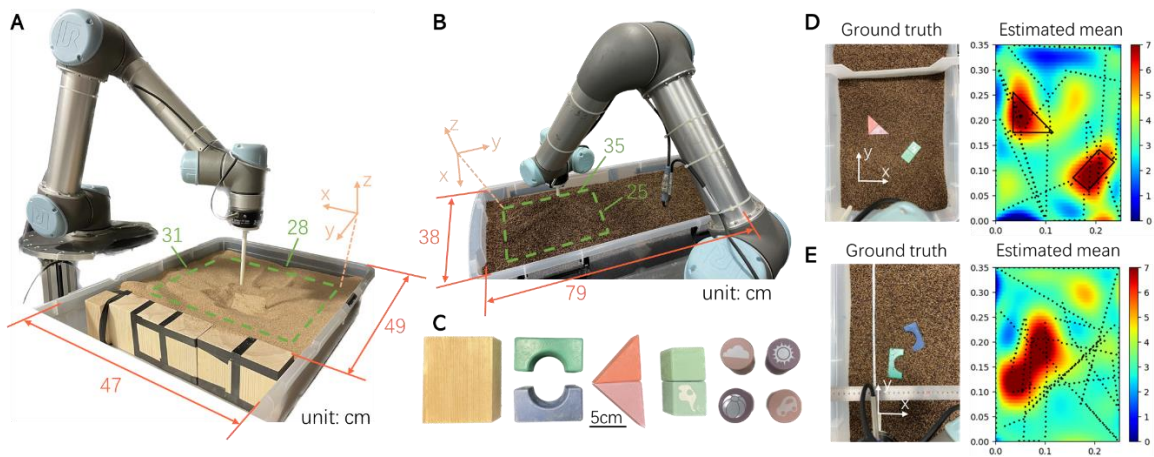

**Fig. S5. BOA-guided pre-touch exploration strategy.** (**A**) Experimental setup filled with sand. (**B**) Experimental setup filled with cassia seed. (**C**) Buried objects used in this work. (**D**) (**E**) Estimated distribution of objects beneath cassia seed.

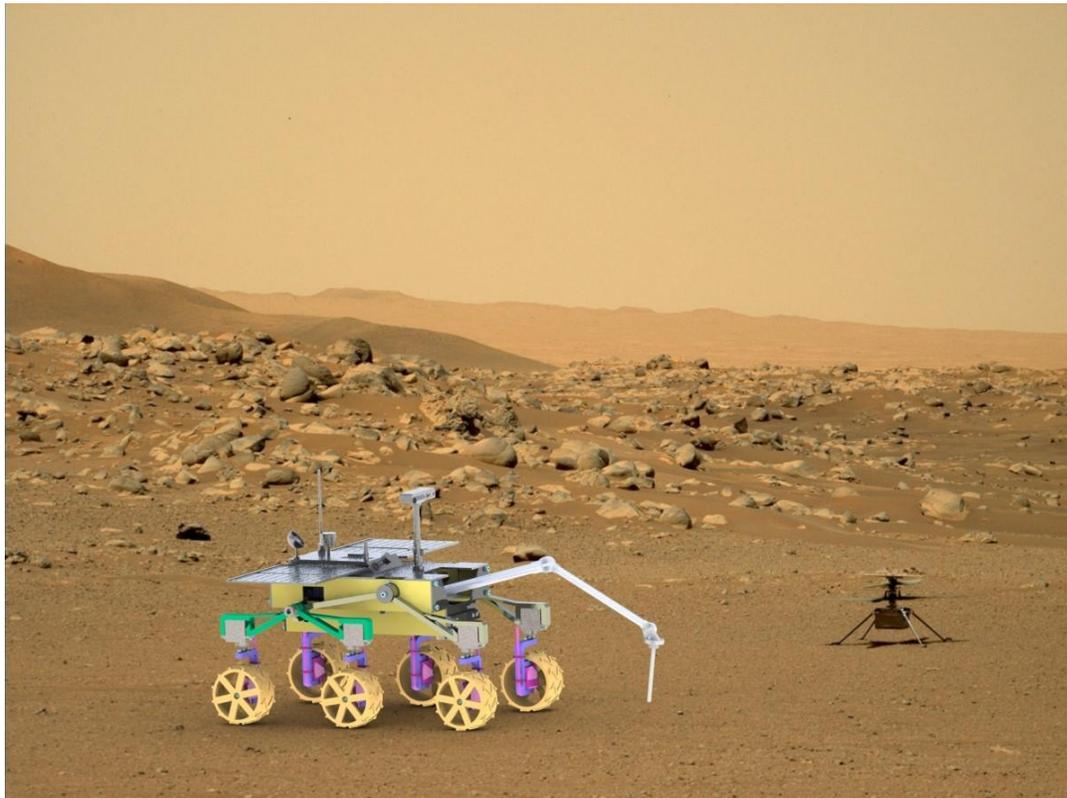

**Fig. S6. A conceptual diagram of unmanned construction robots with GRAINS to detect objects on the granular crust of Mars.** Background: NASA's Ingenuity Mars Helicopter on the Martian surface as seen by the Perseverance rover. (Image credit: NASA/JPL-Caltech/ASU/MSSS)